\documentclass[acmtog,nonacm]{acmart}
\acmSubmissionID{307}

\usepackage{booktabs} 
\usepackage{multirow}

\usepackage[ruled]{algorithm2e} 

\SetAlFnt{\small}
\SetAlCapFnt{\small}
\SetAlCapNameFnt{\small}
\SetAlCapHSkip{0pt}

\acmJournal{TOG}

\begin{document}
\title{Learning a Delighting Prior for Facial Appearance Capture in the Wild}

\author{Yuxuan Han}
\affiliation{%
  \institution{School of Software and BNRist, Tsinghua University}
  \city{Beijing}
  \country{China}
}
\email{hanyx22@mails.tsinghua.edu.cn}
\orcid{0000-0002-2844-5074}

\author{Xin Ming}
\affiliation{%
  \institution{School of Software and BNRist, Tsinghua University}
  \city{Beijing}
  \country{China}
}
\email{1729406968@qq.com}
\orcid{0009-0007-9602-6078}

\author{Tianxiao Li}
\affiliation{%
 \institution{School of Software and BNRist, Tsinghua University}
 \city{Beijing}
 \country{China}
}
\email{tx-li23@mails.tsinghua.edu.cn}
\orcid{0009-0002-5136-6005}

\author{Zhuofan Shen}
\affiliation{%
  \institution{School of Software and BNRist, Tsinghua University}
  \city{Beijing}
  \country{China}
}
\email{czf23@mails.tsinghua.edu.cn}
\orcid{0009-0000-5737-4917}

\author{Qixuan Zhang}
\affiliation{%
 \institution{ShanghaiTech University and Deemos Technology Co., Ltd.}
 \city{Shanghai}
 \country{China}
}
\email{zhangqx1@shanghaitech.edu.cn}
\orcid{0000-0002-4837-7152}

\author{Lan Xu}
\affiliation{%
 \institution{ShanghaiTech University}
 \city{Shanghai}
 \country{China}
}
\email{xulan1@shanghaitech.edu.cn}
\orcid{0000-0002-8807-7787}

\author{Feng Xu}
\affiliation{%
  \institution{School of Software and BNRist, Tsinghua University}
  \city{Beijing}
  \country{China}
}
\email{xufeng2003@gmail.com}
\orcid{0000-0002-0953-1057}

\begin{abstract}

High-quality facial appearance capture has traditionally required costly studio recording. 
Recent works consider an in-the-wild smartphone-based setup; however, their model-based inverse rendering paradigm struggles with the complex disentanglement of reflectance from unknown illumination. 
To bridge this gap, we propose to shift the paradigm into training a powerful delighting network as a prior to constrain the optimization.
We leverage the OLAT dataset and the rendered Light Stage scans for training, and propose Dataset Latent Modulation (DLM) to seamlessly integrate these heterogeneous data sources.
Specifically, by conditioning the core network on learnable source-aware tokens, we decouple dataset-specific styles from physical delighting principles, enabling the emergence of a delighting prior that outperforms existing proprietary models. 
This powerful delighting prior enables a simple and automatic appearance capture pipeline that achieves high-quality reflectance estimation from casual video inputs, outperforming prior arts by a large margin.
Furthermore, we leverage our appearance capture method to transform the multi-view NeRSemble dataset into NeRSemble-Scan, a large-scale collection of 4K-resolution relightable scans. 
By open-sourcing our model and the NeRSemble-Scan dataset, we democratize high-end facial capture and provide a new foundation for the research community to build photorealistic digital humans.

\end{abstract}

\acmJournal{TOG}
\acmYear{2026} \acmVolume{45} \acmNumber{4} \acmArticle{46}
\acmMonth{7} \acmDOI{10.1145/3811303}

%
%


%
%


\begin{teaserfigure}
\centering
    \includegraphics[width=1.0\textwidth]{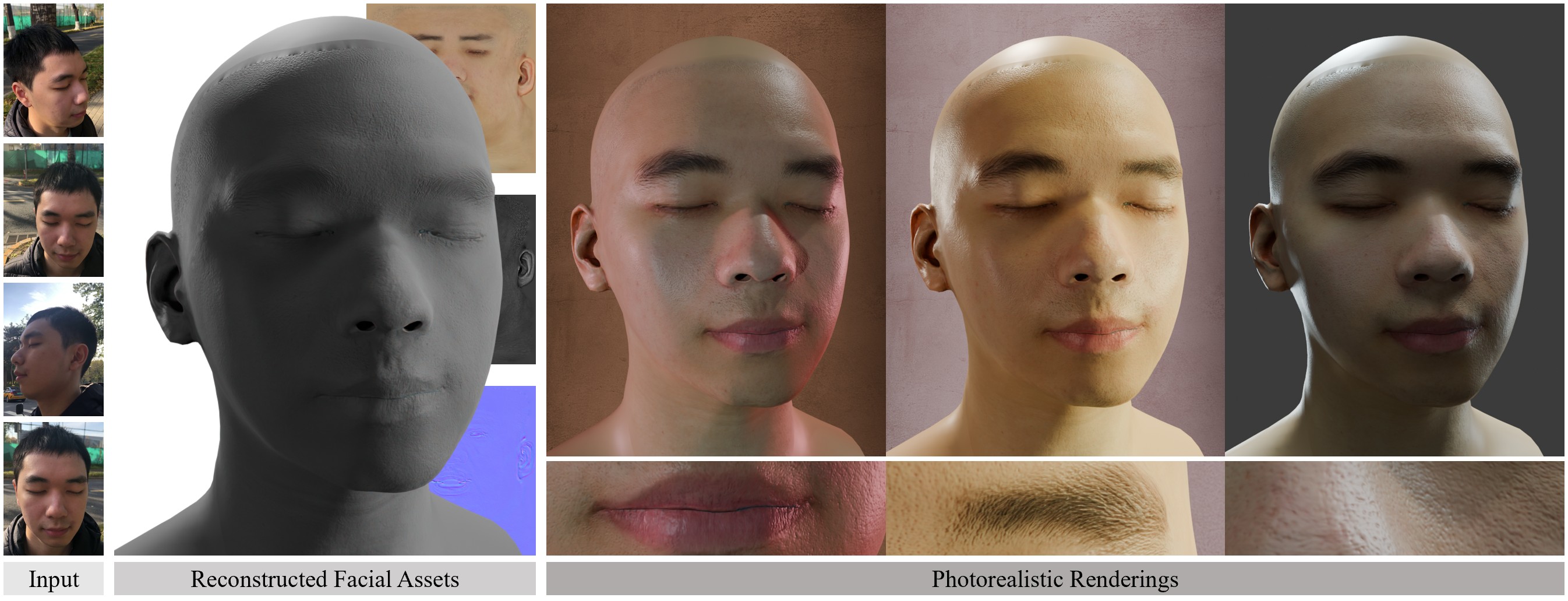}
    \caption{
    Given a smartphone sequence captured in the wild (4 sampled views shown here), our method reconstructs high-quality facial assets, which can be exported to graphics engines like Blender for photo-realistic rendering.
    As shown in the close-up above, our method faithfully reconstructs facial detail patterns around the lip, eyebrow, and cheek.
    }
    \label{fig:teaser}
\end{teaserfigure}

\maketitle

\section{Introduction}
Facial appearance capture lies at the core of digital human creation, enabling applications across visual effects generation, interactive entertainment, and immersive virtual experiences.
Although production-level results have been achieved under delicately controlled lighting using specialized studio systems~\cite{alexander2009digital,alexander2013digital}, such setups are costly and inaccessible to everyday users.
In contrast, capturing facial appearance in the wild, \emph{i.e.}, from casually recorded smartphone videos under unconstrained and unknown illumination, remains fundamentally challenging.
The difficulty stems from faithfully disentangling clean and relightable reflectance maps from complex real-world lighting effects.

Recent years have witnessed a growing interest in facial appearance capture from unconstrained smartphone videos~\cite{xu2024monocular,han2025wildcap,bharadwaj2023flare}.
Early works adopt a model-based inverse rendering framework~\cite{xu2024monocular,bharadwaj2023flare}, which is constrained by oversimplified lighting models, struggling to resolve complex light transport inherent in most in-the-wild scenarios.
To address this, a recent concurrent work, WildCap~\cite{han2025wildcap}, proposes a hybrid approach that performs model-based inverse rendering on top of delighted images predicted by a data-driven method, thereby improving reconstruction quality in various lighting conditions.
However, WildCap necessitates manual efforts to isolate regions where the delighted images exhibit artifacts, thereby hindering its scalability for large-scale applications.
In addition, WildCap relies on a specialized region-based inverse rendering to clean these artifacts, which also increases system complexity and results in a discontinuous texture map.  
Crucially, we also observe that a strong delighting prior can effectively regularize model-based inverse rendering; by internalizing accurate light transport into the prior, it is possible to achieve intervention-free delighting and stable per-scene optimization.
Driven by this insight, we shift the focus of in-the-wild facial appearance capture toward learning a robust delighting prior that effectively disentangles diffuse albedo from a single-view in-the-wild face image.

However, training such a powerful delighting prior is non-trivial.
On the one hand, state-of-the-art solutions remain proprietary in both training data and model weights~\cite{kim2024switchlight,yoon2024generative,futschik2023controllable}, which makes it challenging for the research community to reproduce or build upon their performance.
While FaceOLAT~\cite{prao20253dpr}, a recent open-source dataset captured by a Light Stage in the One-Light-At-a-Time (OLAT) mode~\cite{debevec2000acquiring}, has attempted to bridge this gap, its utility for training a powerful delighting prior is compromised by spatial blurring stemming from imperfect OLAT frame alignment and the lack of physically correct diffuse albedo for supervision.
On the other hand, even the leading proprietary model, SwitchLight~\cite{kim2024switchlight}, struggles with complex lighting and leaves noticeable baking artifacts in predictions, which ultimately necessitate manual interventions and specialized inverse rendering in WildCap~\cite{han2025wildcap}.

In this paper, we propose OpenDelight, a powerful and open-source delighting prior for facial appearance capture in the wild.
On the data front, to address the inherent quality limitations of FaceOLAT~\cite{prao20253dpr}, we first collect a dataset captured by a Light Stage in the material mode~\cite{ghosh2011multiview}, comprising relightable scans of 120 subjects.
Our scan dataset and FaceOLAT are highly complementary: scan-based renderings provide the necessary sharpness and physically correct diffuse albedo, while OLAT-integrated images contribute to closing the gap between synthetic and real-world images.
However, training a delighting prior that seamlessly inherits the strengths of such heterogeneous data sources remains a challenging task.
If handled naively, the network tends to settle for a compromised solution.

To resolve this, our key insight is to decompose the knowledge within these heterogeneous datasets into shared mechanisms (the high-level foundation of delighting) and dataset-specific styles (color distributions, degree of physical correctness, and spatial sharpness).
Building on this insight, we propose a Dataset Latent Modulation (DLM) technique, which employs learnable source-aware tokens as additional conditions to the network.
Given that the parameter count of these tokens is negligible compared to the core network, simply end-to-end training encourages a natural emergence of disentanglement: the tokens absorb the styles of each dataset, while the core network distills the principles of facial delighting.
At inference time, we steer the network toward sharp and physically correct predictions by conditioning it on the tokens associated with our scan dataset.
Thanks to the meticulously constructed dataset and effective training techniques, our model achieves significantly better results than the leading proprietary model SwitchLight~\cite{kim2024switchlight}, while keeping the method simple.

With the powerful delighting prior at hand, we further unlock its potential to empower facial appearance capture in the wild.
Given smartphone-captured multi-view images, we utilize OpenDelight to estimate diffuse albedo across views, which are then fused into a unified texture.
This clean texture base enables performing conventional model-based inverse rendering within a patch-level diffusion prior~\cite{han2025dora,han2025wildcap} in a more stable way.
Compared to WildCap, our method not only eliminates the need for manual intervention to correct network artifacts but also removes its complex and specialized inverse rendering process, while achieving on par or better quality, demonstrating that a superior delighting prior can fundamentally simplify the appearance capture problem.

Finally, leveraging our automatic appearance capture method, we transform an existing multi-view face dataset, NeRSemble~\cite{kirschstein2023nersemble}, into a large-scale collection of 4K-resolution Light Stage scans, which we term NeRSemble-Scan. 
Remarkably, we demonstrate that OpenDelight, when retrained solely on NeRSemble-Scan and FaceOLAT, achieves comparable delighting performance to the model trained on our private Light Stage scan dataset and FaceOLAT.
By open-sourcing our code and the NeRSemble-Scan dataset\footnote{https://yxuhan.github.io/OpenDelight/index.html}, we aim to lower the barriers to high-end facial capture, delighting, and relighting for the research community and everyday users.

In conclusion, our contributions include:
\begin{itemize}
    \item A fully automatic facial appearance capture system that significantly narrows the quality gap between in-the-wild and controllable recordings.
    \item A powerful delighting prior that outperforms leading proprietary solutions, with a Dataset Latent Modulation technique for effective training on heterogeneous data sources.
    \item An open-sourced, large-scale, 4K-resolution Light Stage scan dataset, that enables the community to perform high-quality facial appearance capture and portrait delighting.
\end{itemize}

\section{Related Works}

\subsection{Facial Appearance Capture}
With the goal of cloning real-world faces into the digital world, facial appearance capture has attracted much attention in recent years~\cite{klehm2015recent}.
Early works consider a high-end setup.
They build professional apparatus~\cite{ma2007rapid,ghosh2011multiview,debevec2000acquiring,riviere2020single,gotardo2018practical,weyrich2006analysis,xu2022improved,lattas2022practical,zhang2022videodriven} in studios for on-site data capture and pursue production-level results~\cite{alexander2009digital,alexander2013digital}.
However, these methods are limited to a few professional users.
In recent years, with the advancement of differentiable rendering and data-driven priors, many works have proposed to democratize this technique for low-cost usage.
A group of works~\cite{lattas2020avatarme,lattas2021avatarme++,lattas2023fitme,Paraperas_2023_ICCV,galanakis2025fitdiff,dib2021practical,Dib_2024_CVPR,smith2020morphable,huanglearning,prao20253dpr,prao2024lite2relight,ren2024monocular} reconstruct a relightable scan from a single face image by training on a large-scale dataset captured by the high-end method.
Although user-friendly, the single-view nature hinders these works from achieving production-level quality.
Other works consider a multi-view setup~\cite{xu2024monocular,bharadwaj2023flare,rainer2023neural,zheng2023neuface,li2024uravatar,han2025wildcap,han2024cora,azinovic2023high,wang2023sunstage}.
Despite promising results being achieved by assuming a controllable capture environment via the smartphone flashlight~\cite{han2024cora,han2025dora,azinovic2023high} or the sunlight~\cite{wang2023sunstage}, high-quality appearance capture from in-the-wild smartphone video still remains underexplored. 
In this direction, model-based inverse rendering methods~\cite{bharadwaj2023flare,xu2024monocular} are limited to their oversimplified physical model and unstable optimization process due to the ill-posed nature.
To address this, the concurrent work, WildCap~\cite{han2025wildcap}, proposes a hybrid method that performs model-based inverse rendering on top of delighted images predicted by a data-driven method.
Although it demonstrates improved results, it requires  manual efforts to isolate regions where data-driven predictions exhibit artifacts, thus hindering its scalability for large-scale applications
In this paper, our key insight is to shift the focus of in-the-wild facial appearance capture toward learning a robust delighting prior.
We demonstrate that with a powerful delighting prior at hand, the appearance capture problem becomes much simpler.
As a fully automatic method, we obtain on par or better results compared to the concurrent work with manual intervention, \emph{i.e.}, WildCap.

\subsection{Portrait Delighting Network}
Given an in-the-wild face image, previous works~\cite{wang2020single,pandey2021total,yeh2022learning,kim2024switchlight,Weir2022DeepPD} train a neural network to predict the diffuse albedo image from it.
Another very relevant task is portrait shadow editing~\cite{yoon2024generative,ponglertnapakorn2023difareli,ponglertnapakorn2025difarelidiffusionfacerelighting,futschik2023controllable,zhang2020portrait}, where a network is trained to soften or strengthen the shadow. 
This portrait delighting and shadow editing network has numerous applications, including image-based relighting~\cite{wang2020single,pandey2021total,yeh2022learning} and enhancing other face-perceiving tasks~\cite{Weir2022DeepPD,yoon2024generative}, such as face parsing.
The main challenge in this field is the closed-source ecosystem.
As the training data is captured by a Light Stage~\cite{debevec2000acquiring,ghosh2011multiview} or other controllable environments~\cite{gross2010multi}, which are typically expensive, state-of-the-art solutions remain proprietary in both data and model weights.
While some recent concurrent works~\cite{prao20253dpr,Chen2025POLARAP} have attempted to bridge this gap by releasing their OLAT dataset, there are no existing open-source delighting methods that achieve a comparable performance to the proprietary model.
In this paper, we propose to learn a powerful delighting network to enhance facial appearance capture.
However, unlike the generic methods mentioned above, our delighting network concentrates exclusively on the facial skin region.
By leveraging our Dataset Latent Modulation technique for joint training on the OLAT and the rendered scan dataset, we produce delighting results with negligible lighting residuals that the leading proprietary model (\emph{i.e.}, SwitchLight~\cite{kim2024switchlight}) cannot achieve.
Our code and the training data will also be open-sourced to foster future research.

\subsection{Training Across Face Datasets}
Existing face datasets are heterogeneous, including large-scale single-view datasets, \emph{e.g.}, FFHQ~\cite{karras2019style}, large-scale rendered datasets, \emph{e.g.}, Cafca~\cite{buehler2024cafca}, and small multi-view studio-captured datasets, \emph{e.g.}, NeRSemble~\cite{kirschstein2023nersemble}.
Recent works attempt to train over multiple datasets to inherit generalization ability and quality from different data sources.
Early methods adopt a pretrain-then-finetune strategy~\cite{han2023learning,yoon2024generative}.
However, this inevitably leads to catastrophic forgetting and requires a complex multi-stage training regime.
More recently, some concurrent works~\cite{kirschstein2025flexavatar,Xu2026RelightAnyoneAG} propose simultaneously training on heterogeneous data sources via a learnable latent code to help unified learning.
Despite these works sharing the same spirit as our method at the abstract technical level, the motivation is quite different. 
Concurrent works focus on avatar learning, where they use a learnable latent code to enforce a shared latent space with the same color style~\cite{Xu2026RelightAnyoneAG} or improving 3D-awareness~\cite{kirschstein2025flexavatar}.
Differently, our goal is to train a delighting network where we adopt source-aware tokens to disentangle dataset-specific style from the foundation of delighting mechanism.
In addition, our data sources have a larger domain gap in terms of realism, where we train on a synthetic rendered scan dataset and an OLAT dataset, while concurrent works use different real-face datasets for training.

\section{Overview}
Our method takes a casual smartphone video captured in unconstrained environments as input.
From this source, we aim to reconstruct high-quality and relightable facial assets, comprising a base mesh and a set of reflectance maps, including diffuse albedo, specular albedo, and normal.
However, directly solving for these assets via model-based inverse rendering is notoriously ill-posed~\cite{xu2024monocular,bharadwaj2023flare}, and often suffers from unstable optimization and baked-in shadows in results.

To address this, our key observation is that a powerful data-driven delighting prior can be synergistically integrated with optimization-based frameworks to effectively constrain the inverse rendering process.
Following this insight, we first introduce a powerful delighting prior, OpenDelight, in Section~\ref{sec:delight_pior}, which estimates the diffuse albedo from a single face image.
We then propose an automatic facial appearance capture pipeline in Section~\ref{sec:inv_render}.
In Section~\ref{sec:nsb}, we apply our appearance capture method to transfer NeRSemble~\cite{kirschstein2023nersemble} into an open-source Light Stage scan dataset to foster future research.

\section{Learning a Delighting Prior}\label{sec:delight_pior}
Our goal is to develop a powerful delighting prior tailored for the facial appearance capture task.
To this end, we first construct a large-scale dataset of training pairs in Section~\ref{sec:delight_prior:data}, consisting of face images and their corresponding diffuse albedo images.
Then, we train a neural network in Section~\ref{sec:delight_prior:train} to learn the mapping from images with complex lighting effects to their diffuse albedo.

\subsection{Dataset}\label{sec:delight_prior:data}

We leverage two highly complementary data sources for training, \emph{i.e.}, an OLAT dataset and a synthetic dataset rendered from Light Stage scans, which corresponds to the two capture modes of the Light Stage~\cite{debevec2000acquiring,ghosh2011multiview}.
Below, we first detail the OLAT dataset in Section~\ref{sec:delight_pior:data:olat} and the Light Stage scan dataset in Section~\ref{sec:delight_pior:data:syn}.
Then, we introduce our data engine to convert these datasets into the training format in Section~\ref{sec:delight_pior:data:data_engine}.

\begin{figure}[t]
    \centering
    \includegraphics[width=0.475\textwidth]{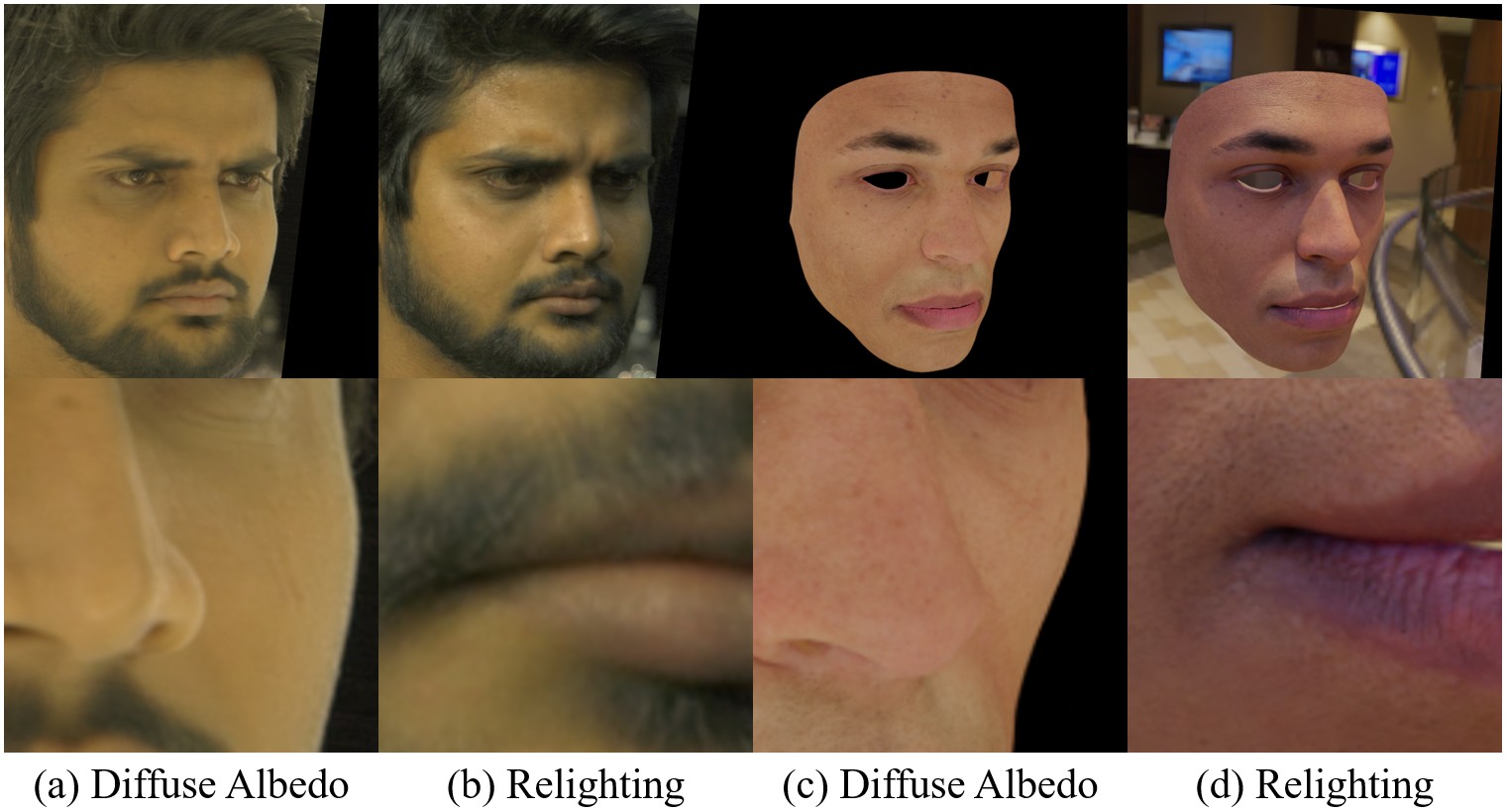}
    \caption{
    The OLAT dataset and the rendered Light Stage scan dataset are complementary data sources for delighting network training. 
    We show (a) a diffuse albedo image (OLAT rendered with uniform lighting) and (b) a relit image from the FaceOLAT dataset~\cite{prao20253dpr}, followed by (c) a diffuse albedo image and a relit image from our rendered scan dataset.
    The rendered scan dataset provides more physically correct diffuse albedo and a relit image with sharper details, while the OLAT dataset has a smaller domain gap relative to real-world images.
    }
    \label{Fig:dataset}
\end{figure}

\subsubsection{OLAT Dataset}\label{sec:delight_pior:data:olat}
We utilize FaceOLAT~\cite{prao20253dpr}, which is the largest publicly available portrait OLAT dataset to date.
It consists of 139 subjects, each captured from 40 viewpoints under 331 individual lighting directions; see more details in their paper~\cite{prao20253dpr}.
We successfully processed 134 subjects.
We use 124 for training and 10 for evaluation.
By leveraging these OLAT data, we can synthesize photo-realistic face images under arbitrary novel illuminations via linear combination.
Following previous works~\cite{wang2020single,pandey2021total}, we approximate the diffuse albedo as images rendered with uniform lighting.

However, training solely on this dataset is insufficient due to its inherent quality issue, as illustrated in Figure~\ref{Fig:dataset}. 
Firstly, since the OLAT images exhibit subtle misalignments that optical flow cannot fully resolve, the integrated results tend to be blurred.
Secondly, because FaceOLAT does not provide polarized captures, treating uniform lighting as diffuse albedo leads to specular baking artifacts, particularly at grazing angles.
In addition, we observe that the trained neural network will inevitably inherit these dataset-specific artifacts, resulting in suboptimal performance.

To circumvent these limitations, we incorporate a synthetic dataset rendered from Light Stage scans for joint training, as it provides high-quality and physically correct supervision signals. 

\subsubsection{Light Stage Scan Dataset}\label{sec:delight_pior:data:syn}
We purchased 120 Light Stage scans from an online store, each with 4K-resolution diffuse albedo, specular albedo, and normal map.
The dataset contains 15 Asians (7 males and 8 females), 23 African Americans (12 males and 11 females), and 82 Caucasians (40 males and 42 females).
Compared to the FaceOLAT dataset, this dataset provides sharper rendered images and more physically correct diffuse albedo, as shown in Figure~\ref{Fig:dataset}.
However, as it is rendered via a graphics engine, this dataset has a wider domain gap relative to real-world images compared to the OLAT dataset.

\begin{figure}[t]
    \centering
    \includegraphics[width=0.475\textwidth]{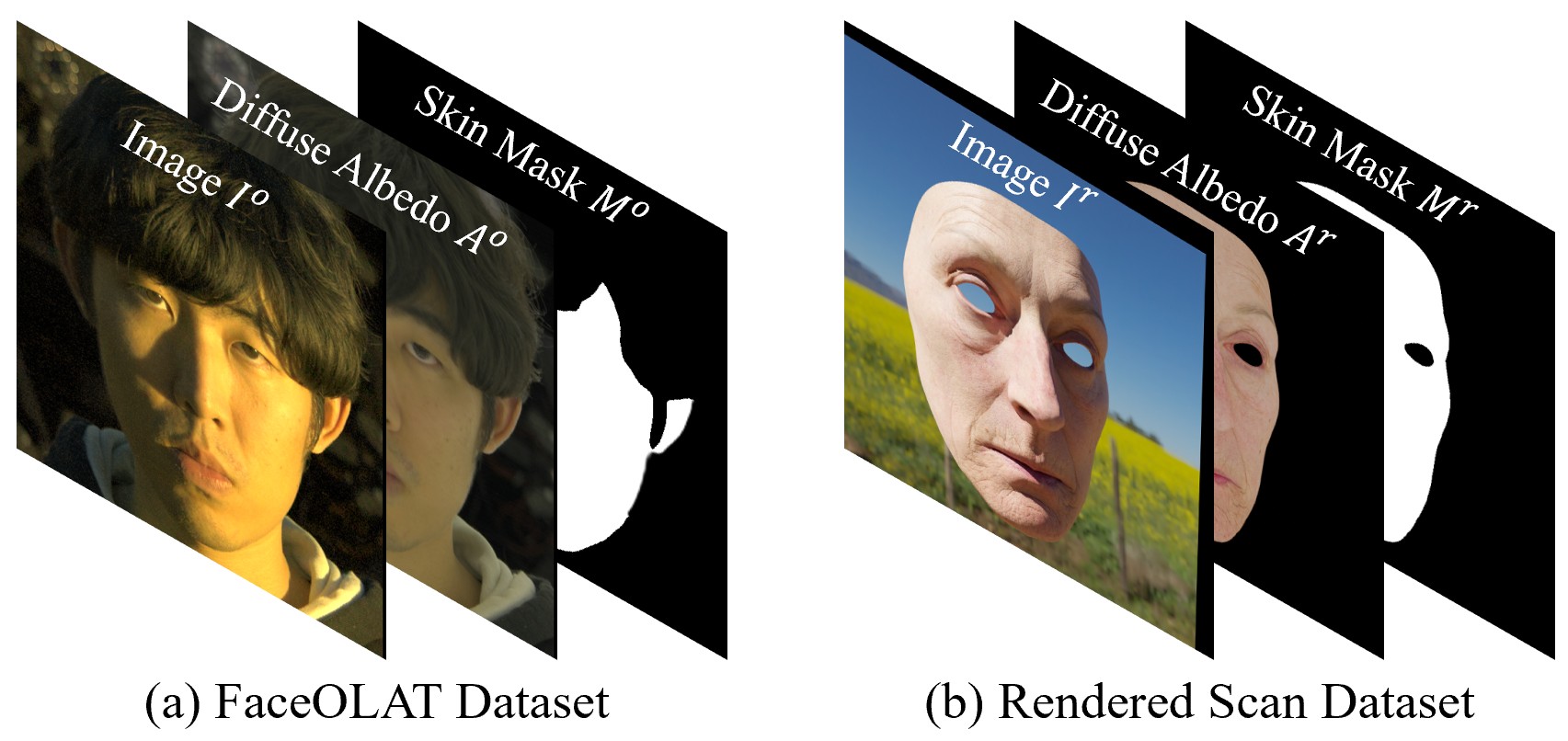}
    \caption{
    Example training pairs generated from our mixing datasets.
    }
    \label{Fig:train_pair}
\end{figure}

\begin{figure*}[t]
    \centering
    \includegraphics[width=\textwidth]{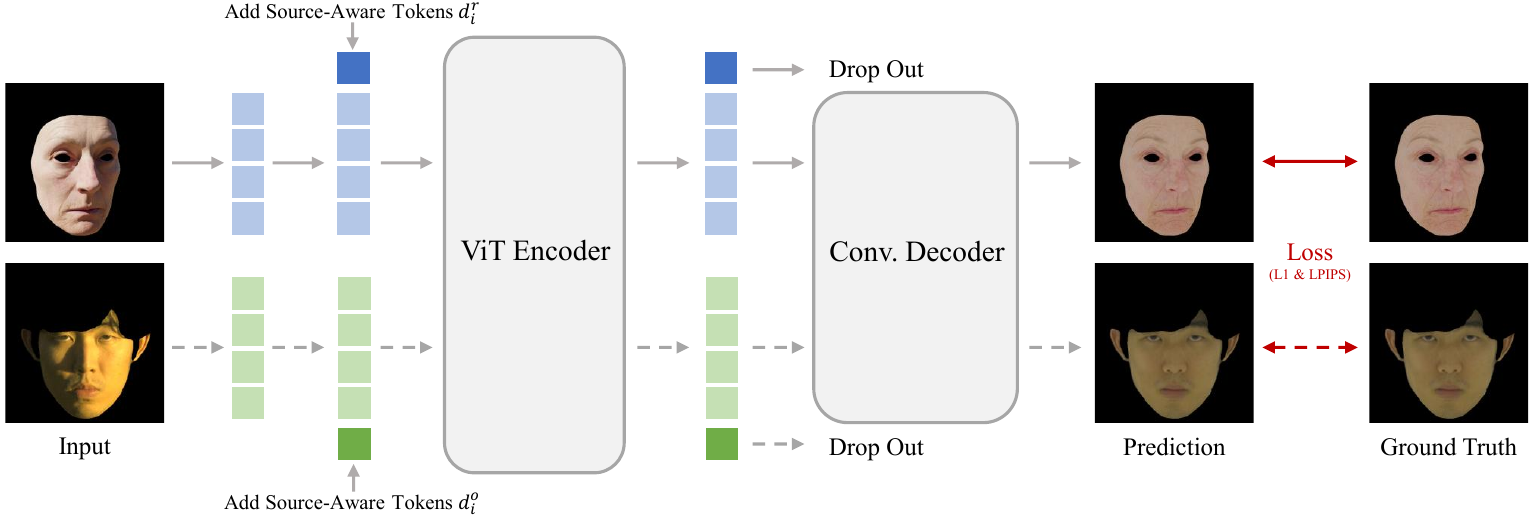}
    \caption{
    Architecture and training pipeline of our base delighting network $F_{base}$. 
    To address the challenge of training on heterogeneous data sources, we propose a Dataset Latent Modulation technique, where we input additional source-aware tokens into the ViT encoder.
    These tokens encourage a natural emergence of disentanglement: the tokens absorb the styles of each dataset, while the core network distills the essential principles of facial delighting.
    }
    \label{Fig:train}
\end{figure*}

\subsubsection{Data Engine}\label{sec:delight_pior:data:data_engine}
To synthesize image-albedo pairs for training, we curate a High Dynamic Range Imaging (HDRI) environment map dataset from PolyHaven, consisting of 473 maps. 
We use 373 maps for training and 100 for evaluation.
We further classify the HDRIs according to their frequency~\cite{sun2020light} and increase the sampling ratio of high-frequency lighting to render more challenging training data; this leads to slightly better delighting results compared to uniformly sampling.
We adopt random rotation augmentation to the HDRIs before rendering.
For the FaceOLAT dataset, we utilize the official scripts to render images under the sampled HDRI lighting, producing approximately 20K image-albedo pairs.
For the Light Stage scan dataset, we render 20K image-albedo pairs using Blender.

To fill the completeness gap, previous works typically collect a large-scale 3D assets library including various clothes and hairstyles, so that the rendered scans are as complete as the real face images~\cite{saleh2025david,yoon2024generative}.
However, since our model is tailored for facial appearance capture, such labourious asset creation is unnecessary.
Instead, we generate skin masks for both datasets and learn a delighting prior over only the skin region.
Following the standard protocol in facial analysis~\cite{zheng2022farl,bulat2017far}, we apply a landmark-based alignment to all the images.
See Figure~\ref{Fig:train_pair} for an example of training pairs from both datasets.
We denote the image, diffuse albedo, and skin mask as $I^*,A^*,M^*$, respectively. 
Here, $*$ can be $o$ or $r$, indicating whether it is from the OLAT dataset or the rendered dataset.

To foster reproducibility, our data engine will be made publicly available, including the HDRI dataset and the rendering pipeline.

\subsection{Network}\label{sec:delight_prior:train}
Given the OLAT dataset and the rendered dataset, our goal is to train a network over these complementary sources to enable generalizable and high-quality delighting.

Following recent advancements~\cite{khirodkar2024sapiens}, we employ a Vision Transformer (ViT)-based encoder ${E}$~\cite{dosovitskiy2020vit} coupled with a lightweight convolutional decoder ${D}$ as the base delighting network ${F}_{base}$. 
We initialize the ViT encoder with Masked Autoencoder (MAE) pre-trained weights~\cite{MaskedAutoencoders2021} while the convolutional head is randomly initialized.

To train ${F}_{base}$, a straightforward strategy is to pre-train it on the OLAT dataset and subsequently fine-tune it on the rendered dataset.
However, we observe that the network loses its robustness to real images due to catastrophic forgetting.
Alternatively, we can jointly train ${F}_{base}$ on both datasets. 
While this mitigates forgetting, it introduces instability: during inference, the network unpredictably switches between the distribution of OLAT and the rendered scan, as shown in Figure~\ref{Fig:exp:eval_dlm}.
This leads to multi-view inconsistency and hinders downstream facial appearance capture.
To resolve these limitations, we propose the Dataset Latent Modulation (DLM) technique.

\subsubsection{Dataset Latent Modulation}\label{sec:delight_prior:train:dlm}
Our key insight is to disentangle the knowledge within these disparate datasets into shared mechanisms (the high-level foundation of delighting) and dataset-specific styles (color distributions, degree of physical correctness, and spatial sharpness).
In this way, we can explicitly control the style of predictions at inference time to ensure view-consistency.

Specifically, as shown in Figure~\ref{Fig:train}, we optimize a set of source-aware tokens $\{d_i^*\}_{i=1}^K$ during training; similarly, $*$ can be $o$ or $r$, corresponding to the OLAT dataset and the rendered dataset.
Given a face image $I^*$ as input, we first tokenize the masked version of it to obtain a series of image tokens $\{f_i\}_{i=1}^N$:
\begin{equation}
    [f_1,f_2,...,f_N] = {\rm Tokenize}(I^*\cdot M^*)
\end{equation}
We then append the source-aware tokens into the image token list and input it to the ViT encoder ${E}$ to obtain a list of encoded tokens:
\begin{equation}
    [\tilde{f_1},\tilde{f_2},...,\tilde{f_N} | \tilde{d_1^*}, \tilde{d_2^*},..., \tilde{d_K^*}] = {E}([f_1,f_2,...,f_N | d_1^*, d_2^*, ..., d_K^*])
\end{equation}
Since the dataset-specific style has already injected into the image tokens, we drop out $\{\tilde{d_i^*}\}_{i=1}^K$ and send the encoded image tokens $\{\tilde{f_i}\}_{i=1}^N$ into the convolutional decoder ${D}$ to produce the predicted diffuse albedo image $\hat{A^*}$:
\begin{align}
    \tilde{f} &= {\rm Unpatchify}([\tilde{f_1},\tilde{f_2},...,\tilde{f_N}]) \\
    \hat{A^*} &= {D}(\tilde{f})
\end{align}
During training, we randomly sample pairs from the OLAT dataset or the rendered dataset.
We apply a combination of L1 loss and LPIPS loss~\cite{zhang2018perceptual} between $\hat{A^*}$ and its ground truth $A^*$.
As the parameter count of these source-aware tokens is negligible compared to the core network, simply end-to-end training encourages a natural emergence of disentanglement: the tokens absorb the styles of each dataset, while the core network distills the essential principles of facial delighting.

At inference time, given an in-the-wild image $I_{raw}$, we first apply a landmark-based alignment and mask out its non-skin region to obtain $I$.
We then send $I$ and the source-aware tokens of the rendered dataset, \emph{i.e.}, $\{d_i^r\}_{i=1}^K$, to the delighting network ${F}_{base}$ to produce the predicted diffuse albedo $\hat{A}$:
\begin{equation}
    \hat{A} = {F}_{base}(I,\{d_i^r\}_{i=1}^K)
\end{equation}
This way, we steer the network toward sharp and physically correct predictions while inheriting strong generalization ability from the OLAT training dataset.

\subsubsection{Detail Enhancement}\label{sec:delight_prior:train:detail_enhance}
Despite strong delighting performance, we observe that the absence of skip connections in our ViT-based architecture leads to a loss of fine-grained details. 
To address this, we introduce a UNet-based detail enhancement network ${F}_{detail}$. 
Specifically, ${F}_{detail}$ takes the concatenation of the original image $I$ and the predicted diffuse albedo $\hat{A}$ as inputs, producing a refined output $\hat{A_{d}}$ with enhanced details:
\begin{equation}
    \hat{A_{d}} = {F}_{detail}(I,\hat{A})
\end{equation}
We train ${F}_{detail}$ solely on the rendered dataset.
During training, we apply stochastic degradations to the ground truth albedo $A^r$, such as Gaussian noise and Gaussian blur, to generate the degraded version $A^r_{deg}$.
We then train ${F}_{detail}$ to recover $A^r$ from $A^r_{deg}$ and $I$, using the L1 loss and LPIPS loss for supervision.

This detail enhancement strategy is also proposed to refine the predictions of conditional Diffusion Models~\cite{yoon2024generative,liu2025dreamlight}. 
Inspired by its success, here we adapt this approach to enhance our ViT-based architecture.

\section{Prior-Grounded Reconstruction}\label{sec:inv_render}
With the powerful delighting prior at hand, we now demonstrate how it empowers facial appearance capture in the wild.
In the following, we first introduce how we capture and process the data in Section~\ref{sec:inv_render:data}.
Then, we detail our inverse rendering pipeline integrated with the delighting prior in Section~\ref{sec:inv_render:opt}.

\subsection{Data Capture and Processing}\label{sec:inv_render:data}
As illustrated in Figure~\ref{fig:teaser}, our input consists of a casual smartphone video captured by moving around the subject.
The entire capture process spans around 30 seconds, during which non-professional subjects are capable of maintaining a stable pose.

Following WildCap~\cite{han2025wildcap}, we resize the frames to a resolution of 960$\times$720. 
We then perform camera calibration via COLMAP~\cite{schoenberger2016mvs,schoenberger2016sfm} and reconstruct a mesh ${G}$ with ICT topology~\cite{li2020learning} using 2DGS~\cite{Huang2DGS2024} and Wrap3D~\cite{Faceform_Wrap2025}.
We next sample $V=16$ frames $\{I_{raw}^i\}_{i=1}^{V}$ according to sharpness for reflectance optimization.
Please refer to WildCap for more details.

\subsection{Optimization}\label{sec:inv_render:opt}
Our reflectance optimization framework is largely built upon the pipeline established by WildCap~\cite{han2025wildcap} and DoRA~\cite{han2025dora}.
Specifically, we first apply OpenDelight to predict the diffuse albedo for each view.
We denote the prediction as $\{I^i\}_{i=1}^{V}$.
Then, we build a texture map $I_{UV}$ at 1K-resolution from $\{I^i\}_{i=1}^{V}$ using the geometry ${G}$ and camera parameters.

In WildCap, a manual mask is required to specify baking artifacts in $I_{UV}$, and subsequently, these artifacts are explained into local dark lights via a complex optimization process within a patch-level diffusion prior~\cite{han2025dora}.
However, we empirically find that the texture map $I_{UV}$ obtained by our OpenDelight contains negligible baking artifacts, even in complex scenes with high-frequency lighting.
Thus, our optimization becomes much easier: we only optimize a global Spherical Harmonics (SH) lighting~\cite{ramamoorthi2001efficient} along with sampling the patch-level diffusion prior.
Namely, our optimization is identical to the \emph{w/o Texel Grid Lighting} baseline in WildCap.
After sampling all the reflectance maps at 1K-resolution, we adopt a super-resolution (SR) network~\cite{zhang2018rcan} to upsample the maps to 4K; this SR network is trained on paired 1K and 4K Light Stage scan patches.
Please refer to WildCap for more details.


\section{The NeRSemble-Scan Dataset}\label{sec:nsb}
We leverage our facial appearance capture method to transform the NeRSemble~\cite{kirschstein2023nersemble} multi-view face dataset into Light Stage scans.
Specifically, we select 170 subjects without face accessories from the original NeRSemble dataset, including 34 Asians (26 males and 8 females), 130 Caucasians (87 males and 43 females), and 6 people with dark skin (4 males and 2 females)
For each subject, we select one frame with a neutral expression and use the provided 16 multi-view images to run our appearance capture method.
We refer to this processed dataset as NeRSemble-Scan, which will be made publicly available.

Next, we train our delighting network over the NeRSemble-Scan and the FaceOLAT dataset.
We denote this version as OpenDelight*. 
As demonstrated in experiments, despite being a fully open-source model, OpenDelight* still achieves superior performance compared to the leading proprietary model SwitchLight.

\begin{figure*}[t]
    \centering
    \includegraphics[width=0.875\textwidth]{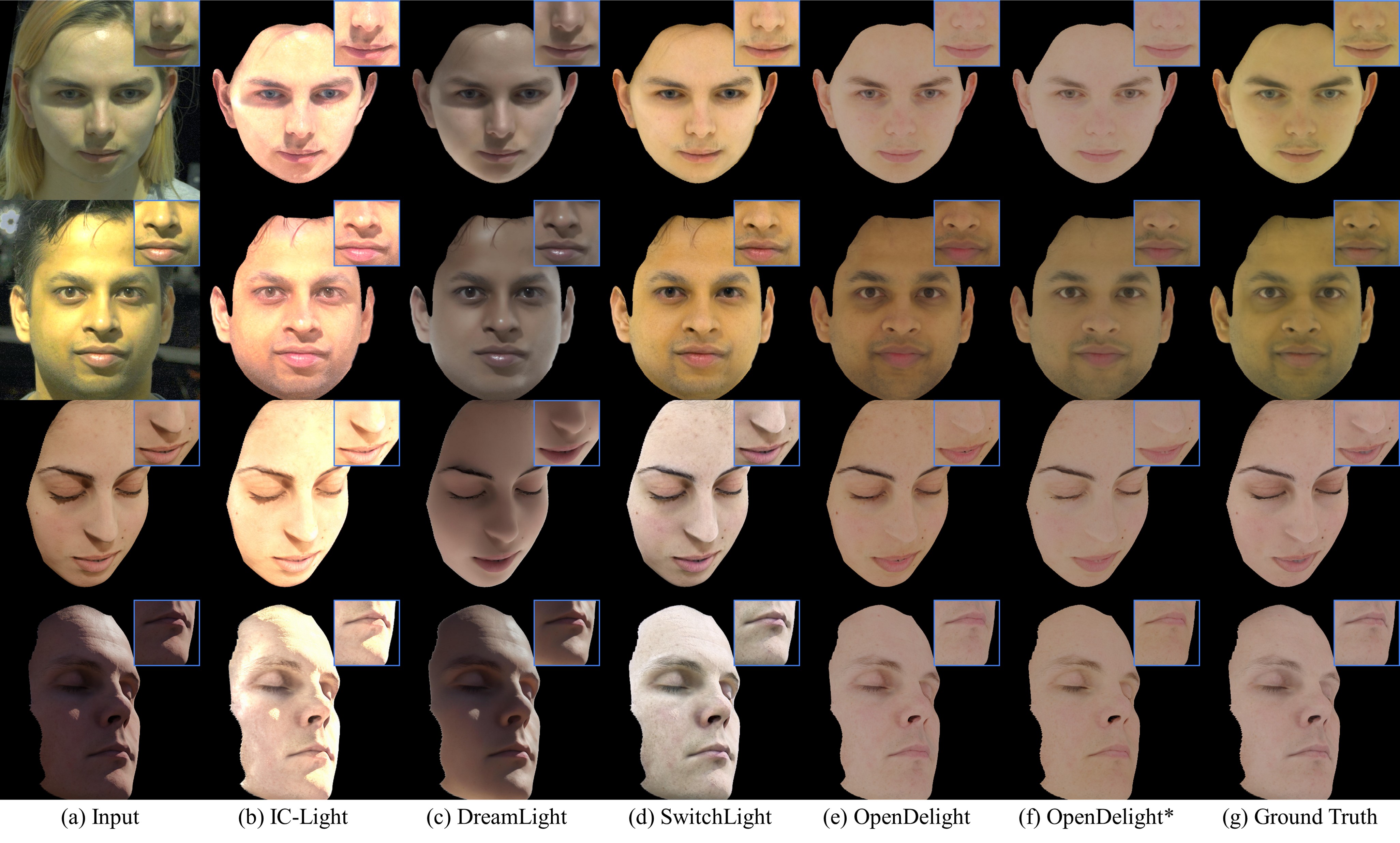}
    \caption{
    Comparison to prior arts in terms of diffuse albedo prediction on the FaceOLAT (the first two rows) and 3DRFE dataset (the last two rows).
    }
    \label{Fig:exp:cmp_testset}
\end{figure*}

\section{Experiments}
In this section, we first introduce implementation details in Section~\ref{sec:exp:imple}.
Then, we evaluate our delighting network (Section~\ref{sec:exp:delight}) and the appearance capture method (Section~\ref{sec:exp:capture}).
Lastly, we discuss the limitations and future works in Section~\ref{sec:exp:discuss}.
We strongly suggest the reader check our supplementary material for more results.

\subsection{Implementation Details}\label{sec:exp:imple}
\subsubsection{Delighting Prior Training}
We adopt the ViT-Base architecture for the encoder ${E}$ in our base delighting network ${F}_{base}$.
We train the network at a resolution of $512\times512$.
The training takes 1.5 days on two 48GB NVIDIA L20 graphics cards with a total batch size of 8.
The learning rate for the ViT encoder ${E}$ and the convolutional decoder ${D}$ are set to $1e-5$ and $1e-4$, respectively.
We set $K=4$ source-aware tokens.
We train the detail enhancement network ${F}_{detail}$ at a resolution of $768\times768$.
The training takes 0.5 days on two 24GB NVIDIA RTX4090 graphics cards with a total batch size of 8.
The learning rate is set to $1e-4$.
At inference time, before sending $\hat{A}$ into ${F}_{detail}$, we first interpolate it to $768\times768$ resolution.
To ensure reproducibility, all the code will be made publicly available.

\subsubsection{Prior-Grounded Reconstruction}
We adopt the same patch-level diffusion prior and the reflectance super-resolution network as WildCap~\cite{han2025wildcap}.
The sampling schedule and Other hyperparameters are kept identical to WildCap.
This stage takes 20 minutes for data capture and processing and 8 minutes for appearance capture on a single 24GB NVIDIA RTX4090 graphics card.
We thank the author of WildCap for sharing their code.

\subsubsection{Test Dataset and Metrics}
We conduct quantitative experiments on the FaceOLAT test set and the rendered scans from an unseen Light Stage dataset, \emph{i.e.}, 3DRFE~\cite{Stratou2012ExploringTE}.
The FaceOLAT test set comprises 10 subjects, while the 3DRFE dataset contains 23 subjects.
We render 20 images for each subject in the FaceOLAT test set, while 10 images for the 3DRFE dataset.
When rendering images, we keep the HDRIs sampled from the evaluation set to ensure the test lightings are unseen during training.

To avoid quantitative evaluation being affected by global color scale ambiguities inherent in different face datasets, we estimate a color transform to align the predictions with the ground truth diffuse albedo before computing metrics.
This color transform includes a channel-wise scale and bias, with a total of 6 parameters, and is estimated on a per-method (including ours and all baselines to ensure fairness), per-image basis\footnote{This color transform is only applied to quantitative results; all qualitative comparisons use the unaligned predictions to show the actual model output.}.
We report Peak Signal-to-Noise Ratio (PSNR), Structural Similarity Index Measure (SSIM), and Learned Perceptual Image Patch Similarity (LPIPS) metrics over the facial region indicated by the skin mask.

\begin{table}[t]
\footnotesize
\caption{Quantitative comparison to prior arts in terms of diffuse albedo prediction on the FaceOLAT and 3DRFE dataset. }
\begin{tabular}{cccclccc}
\hline
\multirow{2}{*}{Methods} & \multicolumn{3}{c}{FaceOLAT}                               &  & \multicolumn{3}{c}{3DRFE}                               \\ \cline{2-4} \cline{6-8} 
                         & \!\!\!PSNR $\uparrow$\!\!\! & \!\!\!SSIM $\uparrow$\!\!\! & \!\!\!LPIPS $\downarrow$\!\!\! & & \!\!\!PSNR $\uparrow$\!\!\! & \!\!\!SSIM $\uparrow$\!\!\! & \!\!\!LPIPS $\downarrow$\!\!\! \\ \hline
IC-Light              & 30.78           & 0.9671          & 0.1573             &  & 31.17           & 0.9568          & 0.1282             \\
DreamLight              & 27.72           & 0.9747          & 0.1847             &  & 28.84           & 0.9627           & 0.2055             \\
SwitchLight                 & \textbf{32.98}           & 0.9800          & 0.1031             &  & 31.08           & 0.9744          & 0.1032             \\ \hline
OpenDelight                 & 32.88           & \textbf{0.9817}          & \textbf{0.0952}             &  & 34.27           & 0.9803          & 0.0649             \\
OpenDelight*                 & 31.93           & 0.9809          & 0.1072             &  & \textbf{35.02}           & \textbf{0.9812}          & \textbf{0.0606}             \\ \hline
\end{tabular}
\label{Tab:quant}
\end{table}

\begin{figure*}[t]
    \centering
    \includegraphics[width=\textwidth]{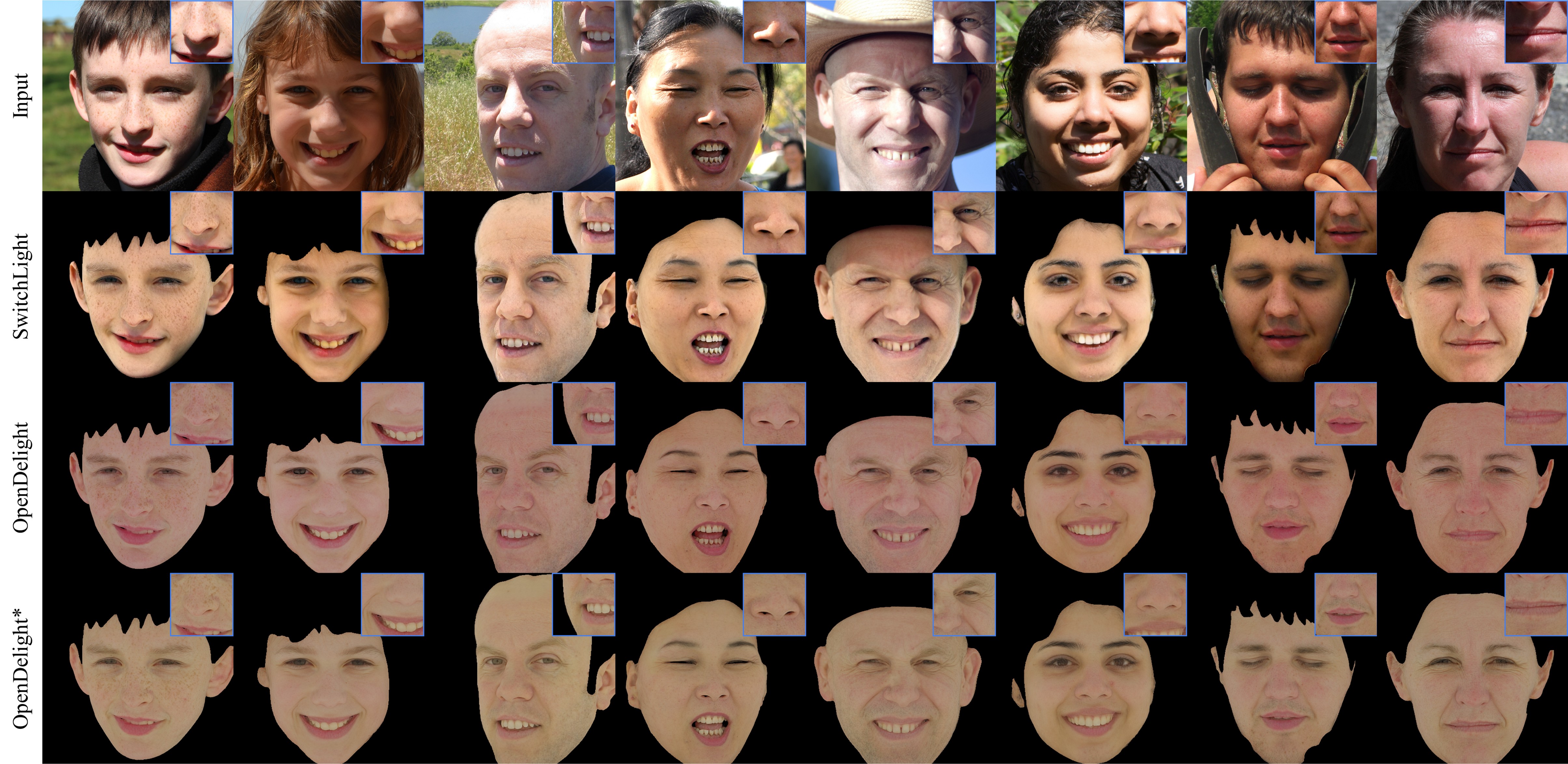}
    \caption{
    Comparison to SwitchLight on in-the-wild face images sampled from the FFHQ dataset in terms of diffuse albedo prediction.
    }
    \label{Fig:exp:cmp_switchlight}
\end{figure*}

\subsection{Evaluation on Delighting Prior}\label{sec:exp:delight}
In the following, we first compare our method to prior arts in Section~\ref{sec:exp:delight:cmp}.
We then evaluate several key design choices, including the importance of training on mixing datasets (Section~\ref{sec:exp:delight:mix_data}), the strategy for mixing training (Section~\ref{sec:exp:delight:mix_train}), and the effectiveness of our detail enhancement network (Section~\ref{sec:exp:delight:unet}).

\subsubsection{Comparisons}\label{sec:exp:delight:cmp}

\paragraph{Baselines}
We extensively compare our method against SwitchLight~\cite{kim2024switchlight}, a leading proprietary model for diffuse albedo prediction. 
We do not compare other existing methods, such as~\citet{pandey2021total} and~\citet{yeh2022learning}, as they are neither publicly accessible nor perform better than SwitchLight.
For the sake of completeness, we also compare methods working on relevant tasks.
Firstly, we include state-of-the-art open-source portrait harmonization models, \emph{i.e.}, IC-Light~\cite{zhang2025scaling} and DreamLight~\cite{liu2025dreamlight}, for a comparison.
Secondly, we compare to GPSR~\cite{yoon2024generative}, a leading proprietary model for portrait shadow removal.
As all the baselines are trained on complete face images, we run them on the complete face image and then compare their results on the facial skin region with our method.
In addition, we include the open-sourced version of our method, OpenDelight*, for comparison, to locate its performance grade.

\paragraph{Comparison with SwitchLight}
SwitchLight~\cite{kim2024switchlight} is a proprietary model trained on a large-scale OLAT dataset.
We first compare our method to it on our test set.
As shown in Table~\ref{Tab:quant} and Figure~\ref{Fig:exp:cmp_testset}, our method demonstrates superior performance on both test sets.
We further conduct qualitative comparisons on in-the-wild subjects sampled from the FFHQ dataset~\cite{karras2019style}.
As shown in Figure~\ref{Fig:exp:cmp_switchlight}, SwitchLight tends to bake lighting effects, particularly in facial regions with hard shadow.
Both OpenDelight and OpenDelight* produce significantly better delighting results than SwitchLight, resulting in a cleaner diffuse albedo.

We believe the performance gain mostly arises from our training data.
In our pilot study, we trained our network on a simplified dataset, where we only adopted low-frequency HDRIs for rendering training pairs. 
In this case, we observe noticeable baking artifacts similar to those seen in SwitchLight.
We emphasize that the open-source nature of our method enables the research community to better study the working mechanism of the delighting network.


\paragraph{Comparison with IC-Light and DreamLight}
IC-Light~\cite{zhang2025scaling} and DreamLight~\cite{liu2025dreamlight} are trained to modify the lighting effects of the portrait image to match the provided background image and/or text prompt.
We adapt IC-Light and DreamLight to the diffuse albedo prediction task by providing a uniform white background image to them.
We have also tried other background images, but find they work best under a uniform white background.
As demonstrated in Table~\ref{Tab:quant} and Figure~\ref{Fig:exp:cmp_testset}, these methods produce inferior diffuse albedo predictions because the network is ambiguous in inferring the lighting effects on the face given only the background image. 
By training specifically on diffuse albedo prediction, our method internalizes the delighting mechanism in network weights and thus achieves better results.

\begin{figure}[t]
    \centering
    \includegraphics[width=0.475\textwidth]{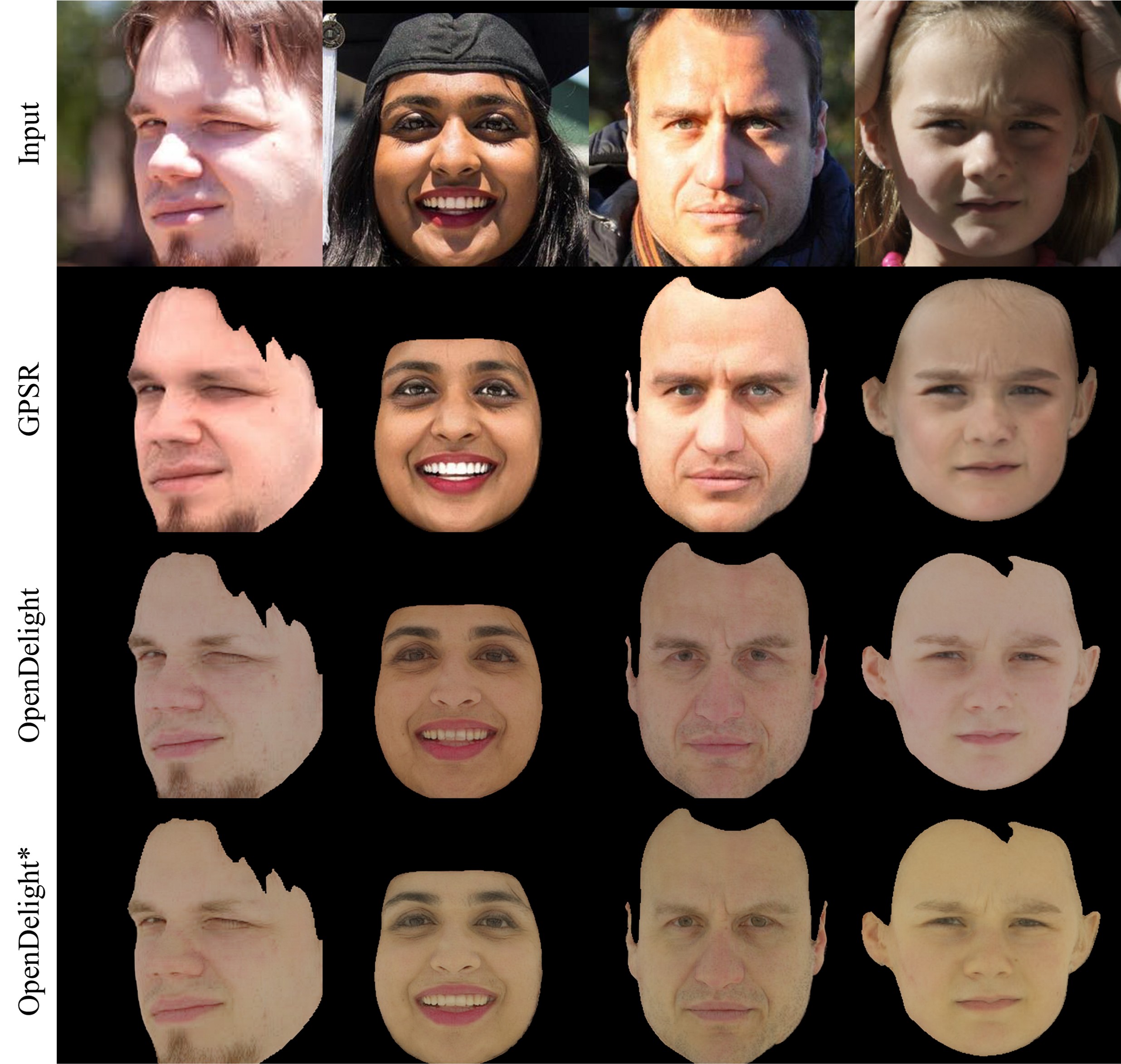}
    \caption{
    Comparison to a portrait shadow removal method, GPSR.
    }
    \label{Fig:exp:cmp_gpsr}
\end{figure}

\paragraph{Comparison with GPSR}
GPSR~\cite{yoon2024generative} is a proprietary model trained on a mixture of an OLAT dataset and a synthetic dataset for portrait shadow removal. 
Since GPSR is not publicly accessible, we conducted a qualitative comparison by running our method on the example images provided in their paper.
As illustrated in Figure~\ref{Fig:exp:cmp_gpsr}, both our method and GPSR can effectively remove complex hard shadows.
Notably, our framework is simpler than GPSR.
While GPSR requires a complex multi-stage training pipeline, our method only requires a single-stage training thanks to the proposed Dataset Latent Modulation technique.
In addition, our inference is substantially more efficient, requiring only 2 forward passes (ViT and UNet) compared to the 50 sampling steps typically required by GPSR's Diffusion Model.

\begin{figure}[t]
    \centering
    \includegraphics[width=0.475\textwidth]{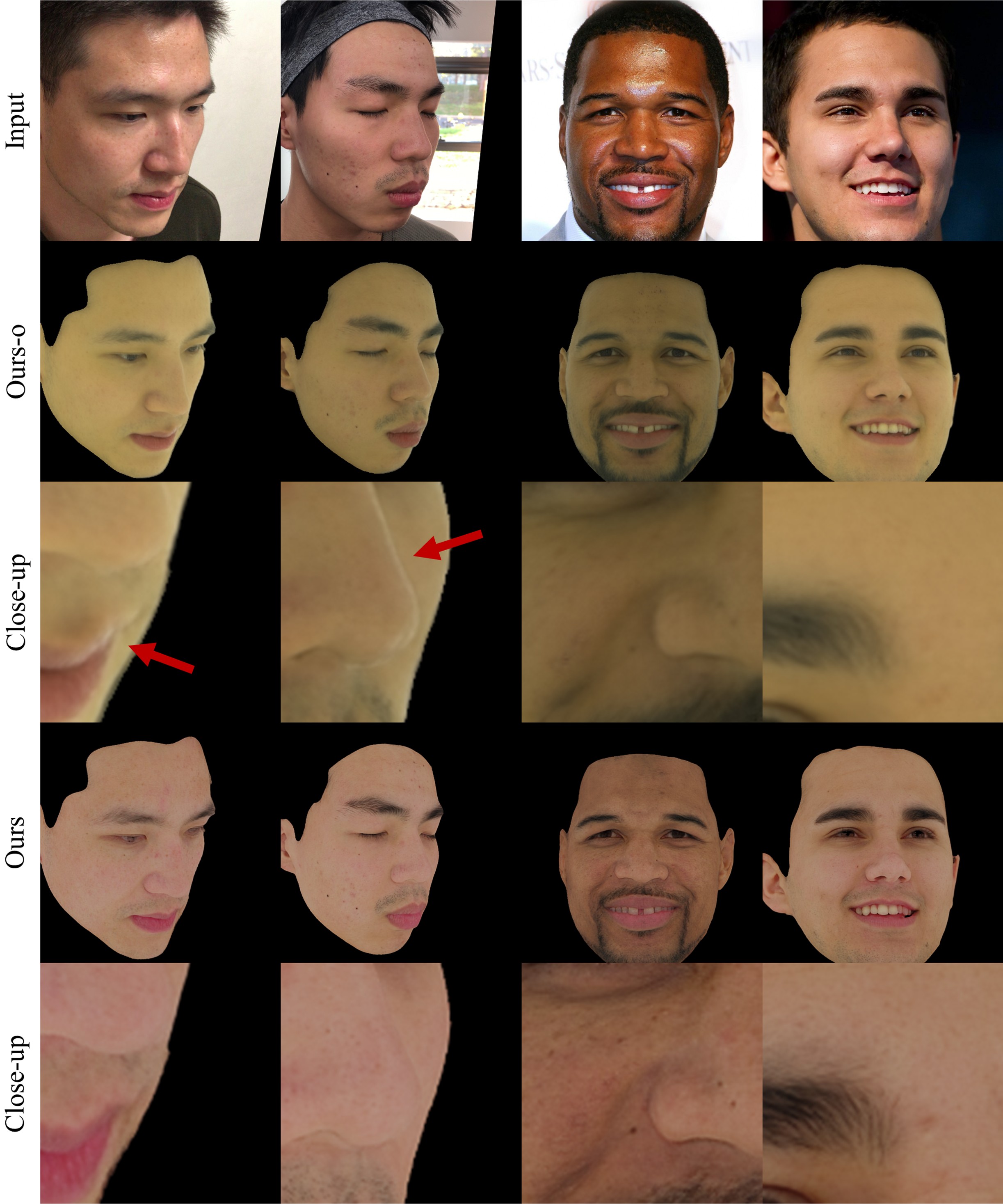}
    \caption{
    Ablation study on mixing dataset training in terms of diffuse albedo prediction.
    We compare a baseline variant \emph{Ours-o}, which is trained solely on the FaceOLAT dataset.
    }
    \label{Fig:exp:eval_faceolat_only}
\end{figure}

\begin{figure}[t]
    \centering
    \includegraphics[width=0.475\textwidth]{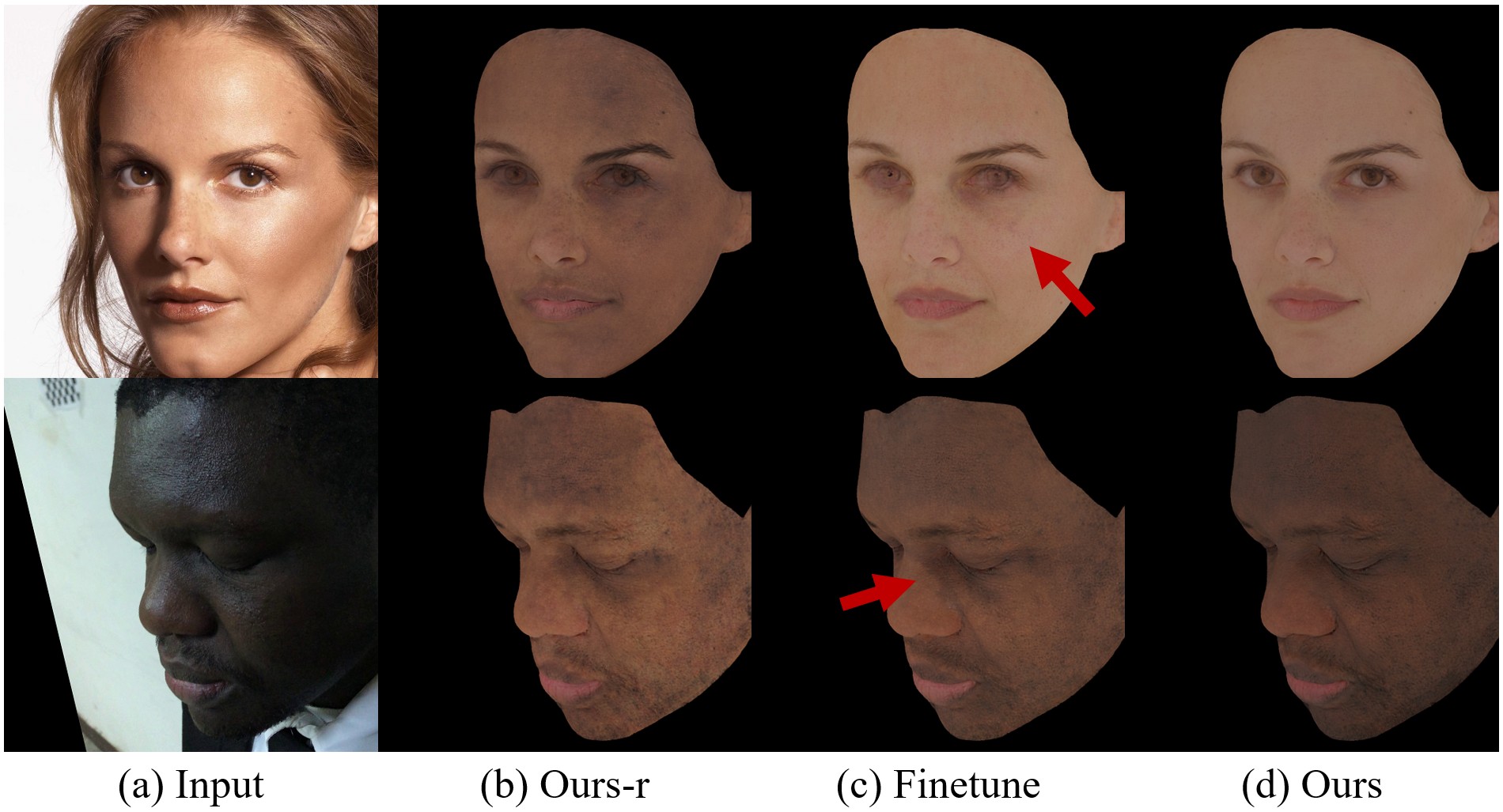}
    \caption{
    Ablation study on mixing dataset training and mixing training strategy in terms of diffuse albedo prediction. 
    We compare two baseline variants: \emph{Ours-r} (trained solely on the rendered scan dataset) and \emph{Finetune} (pretrained on the OLAT dataset and finetuned on the rendered scan dataset).
    }
    \label{Fig:exp:eval_syn_only_finetune}
\end{figure}

\subsubsection{Importance of Mixing Dataset Training}\label{sec:exp:delight:mix_data}
To evaluate the importance of training on the mixing dataset, we introduce two baselines: \emph{i)} \emph{Ours-o}, where we train the network solely on the FaceOLAT dataset, and \emph{ii)}, \emph{Ours-r}, where we train the network solely on the rendered scan dataset. 
To train these two baselines, we remove the source-aware tokens and keep other implementations identical to our full method.
In addition, the detail enhancement network ${F}_{detail}$ in \emph{Ours-o} is trained on the FaceOLAT dataset.

We first compare the baseline trained solely on the FaceOLAT dataset.
As shown in Figure~\ref{Fig:exp:eval_faceolat_only}, \emph{Ours-o} inherits the dataset-specific issues of the FaceOLAT dataset illustrated in Figure~\ref{Fig:dataset}, including spatial blurring and physically incorrect diffuse albedo with specular highlights baking.
Our method effectively resolves these issues by combining a rendered Light Stage scan dataset for mix training, resulting in more physically correct and sharper results. 

We then compare the baseline trained solely on the rendered scan dataset.
As shown in Figure~\ref{Fig:exp:eval_syn_only_finetune}, \emph{Ours-r} sometimes predicts wrong skin tone (the first row) or dirty diffuse albedo (the second row).
That is primarily due to the domain gap between the rendered scan and the real-world images.
By training synergistically with the FaceOLAT dataset, our method achieves significantly better results with plausible skin tone and clean diffuse albedo.

\begin{figure}[t]
    \centering
    \includegraphics[width=0.475\textwidth]{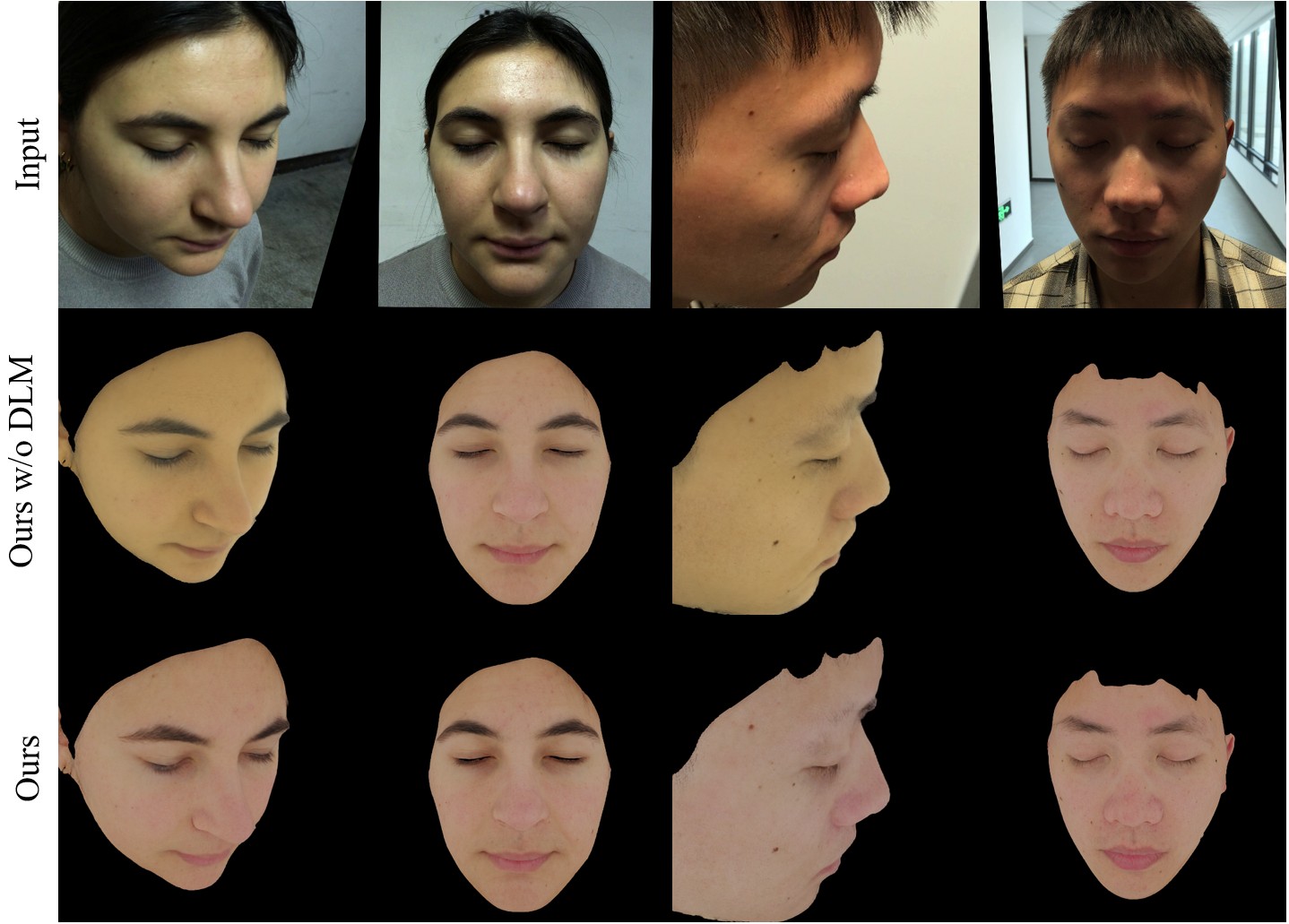}
    \caption{
    Ablation study of the Dataset Latent Modulation (DLM) technique on diffuse albedo prediction. 
    Without DLM, the network unpredictably switches between the distribution of the FaceOLAT dataset and the rendered scan dataset at inference time, leading to severe view inconsistency.
    }
    \label{Fig:exp:eval_dlm}
\end{figure}

\begin{figure*}[t]
    \centering
    \includegraphics[width=0.778\textwidth]{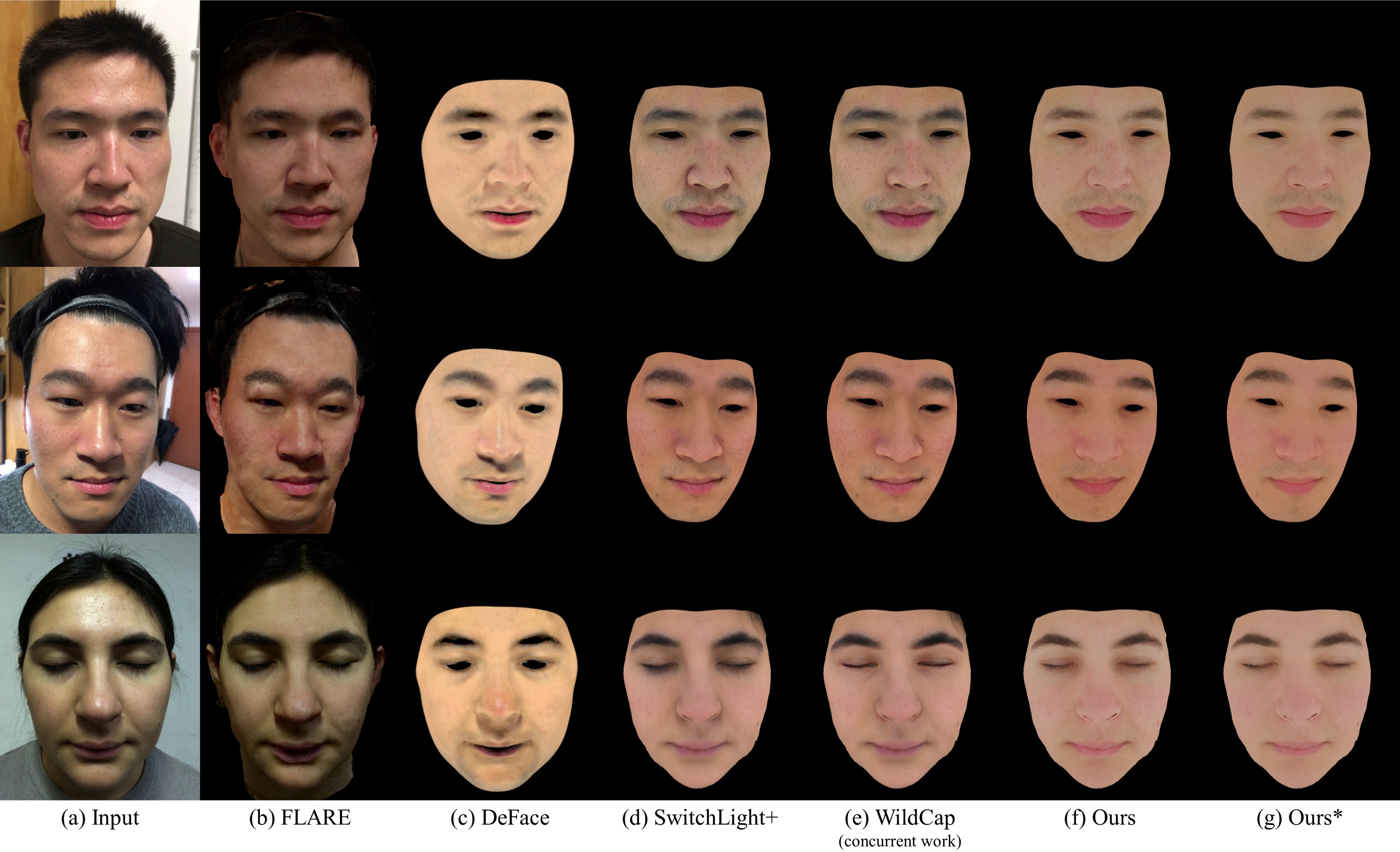}
    \caption{
    Comparison to existing facial appearance capture methods in terms of diffuse albedo reconstruction.
    }
    \label{Fig:exp:cmp_cap}
\end{figure*}

\begin{figure}[t]
    \centering
    \includegraphics[width=0.475\textwidth]{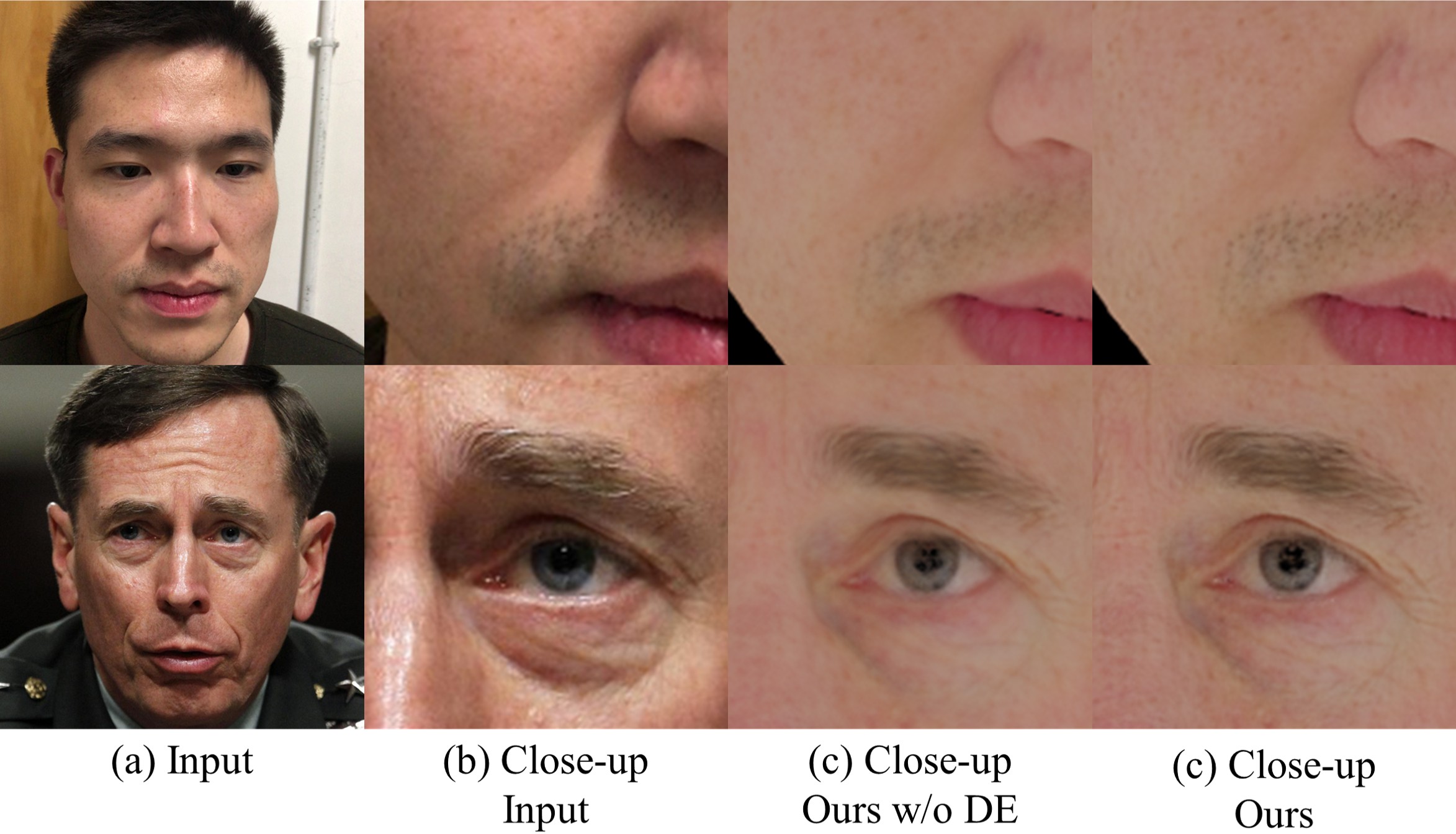}
    \caption{
    Ablation study of the detail enhancement network in diffuse albedo prediction. 
    We compare to a baseline variant \emph{Ours w/o DE}, where we directly use the results of ${F}_{base}$.
    }
    \label{Fig:exp:eval_detail_enhance}
\end{figure}

\subsubsection{Design Choices in Mixing Training Strategy}\label{sec:exp:delight:mix_train}
To evaluate the effectiveness of the proposed Dataset Latent Modulation technique for training on mixing datasets, we conduct two baselines: \emph{i)} \emph{Finetune}, where we pretrain the network on the FaceOLAT dataset and finetune it on the rendered scan dataset, and \emph{ii)}, \emph{Ours w/o DLM}, where we remove the source-aware tokens in our full method.

We first compare to the \emph{Finetune} baseline.
As shown in Figure~\ref{Fig:exp:eval_syn_only_finetune}, the \emph{Finetune} baseline (c) achieves better results than training solely on the rendered scan dataset (b), since it inherits robustness from pretraining on the FaceOLAT dataset.
However, one limitation of \emph{Finetune} is catastrophic forgetting.
This causes the \emph{Finetune} baseline to produce results with compromised quality.
By simultaneously training on both datasets, our method avoids forgetting, produces a cleaner and flatter diffuse albedo (d), especially in regions indicated by the red arrow.

We then compare to the \emph{Ours w/o DLM} baseline in Figure~\ref{Fig:exp:eval_dlm}.
Without the proposed Dataset Latent Modulation technique, \emph{Ours w/o DLM} unpredictably switches between the distribution of two training datasets at inference time, leading to severe view inconsistency.
Our method effectively resolves this issue by explicitly disentangling the dataset-specific style from the fundamental delighting mechanisms.
This way, we ensure view-consistent inference by explicitly indicating the style via the source-aware tokens.

\subsubsection{Importance of Detail Enhancement Network}\label{sec:exp:delight:unet}

To evaluate the importance of the detail enhancement network ${F}_{detail}$, we compare to a baseline variant \emph{Ours w/o DE} where we directly use the results of ${F}_{base}$.
As shown in Figure~\ref{Fig:exp:eval_detail_enhance}, the results from the base delighting network ${F}_{base}$ lack fine-grained details, primarily due to the absence of skip connections in the ViT-based architecture.
By learning a UNet-based enhancer ${F}_{detail}$, we can effectively restore details from the input image, leading to higher-quality results.

\subsection{Evaluation on Appearance Capture}\label{sec:exp:capture}
In the following, we first compare our method to prior arts in Section~\ref{sec:exp:capture:cmp}.
We then evaluate the reconstruction fidelity of our method by comparing the re-rendered image against the real image in Section~\ref{sec:exp:capture:recon_fidelity}.
Lastly, we present more results of our method on diverse ethnic groups in Section~\ref{sec:exp:capture:more_results}.

\subsubsection{Comparisons}\label{sec:exp:capture:cmp}
\paragraph{Baselines}
We first compare existing appearance capture methods that take the same input as our method.
We include optimization-based methods, \emph{i.e.}, FLARE~\cite{bharadwaj2023flare} and DeFace~\cite{huanglearning}, and a state-of-the-art concurrent work, \emph{i.e.}, WildCap~\cite{han2025wildcap}, for comparison. 
To evaluate the importance of our delighting prior, we conduct a baseline variant, \emph{SwitchLight+}, where we replace our OpenDelight with SwitchLight~\cite{kim2024switchlight} while keeping other implementations unchanged.
In addition, we compare to the open-sourced variant, \emph{Ours*}, where we replace OpenDelight with OpenDelight* in our pipeline.
We do not compare to \citet{xu2024monocular} as their method is closed-source.
Beyond involving baselines in the same setup as ours, we compare to DoRA~\cite{han2025dora}, a method that takes a co-located smartphone and flashlight sequence as input.

\paragraph{Comparison with FLARE and DeFace}
FLARE~\cite{bharadwaj2023flare} and DeFace~\cite{huanglearning} are model-based inverse rendering methods that aim to disentangle geometry, reflectance, and lighting from the observed images.
To run their methods, we directly input the captured images $\{I_{raw}^i\}_{i=1}^{V}$ to FLARE while sending a selected frontal view image to DeFace. 
As shown in Figure~\ref{Fig:exp:cmp_cap}, FLARE and DeFace can remove specular highlights to some extent but struggle with hard shadows, which leads to dirty diffuse albedo with baking artifacts.
That is because these model-based methods typically apply an oversimplified physical model, and the disentanglement is inherently ill-posed.
By internalizing complex light transport into a learned delighting prior, our method demonstrates significantly better results. 

\begin{figure}[t]
    \centering
    \includegraphics[width=0.475\textwidth]{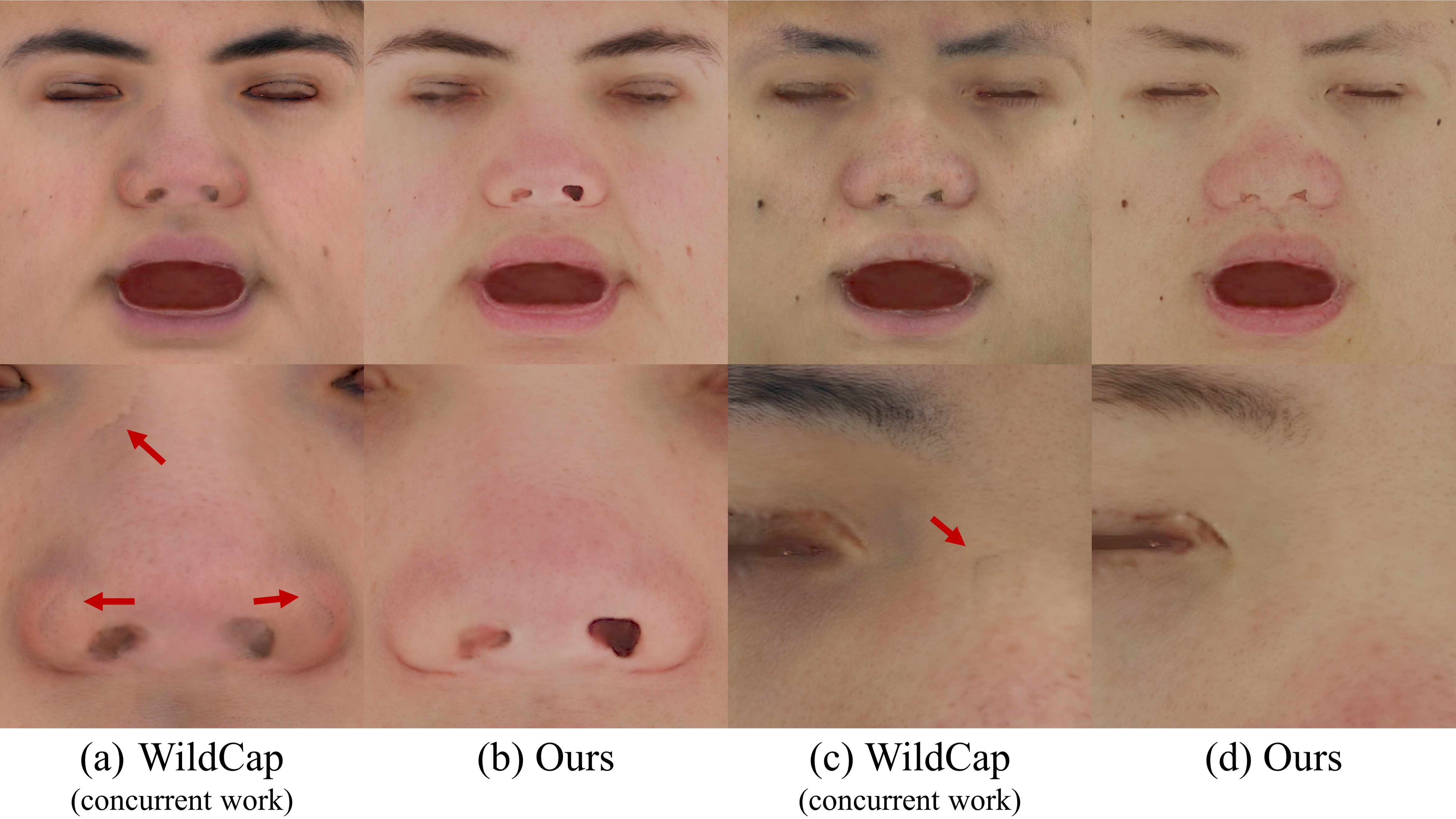}
    \caption{
    Comparison to WildCap on diffuse albedo reconstruction. 
    We show the reconstruction in the first row and close-ups in the second row. 
    }
    \label{Fig:exp:cmp_wildcap}
\end{figure}

\paragraph{Comparison with SwitchLight+ and WildCap}
Both \emph{SwitchLight+} and WildCap~\cite{han2025wildcap} apply a data-driven delighting network, SwitchLight, to remove most lighting effects in the image, and then adopt a model-based inverse rendering pipeline to solve other reflectance maps.
As shown in Figure~\ref{Fig:exp:cmp_cap}, \emph{SwitchLight+} produces better results than model-based methods such as DeFace and FLARE.
However, its results still have artifacts such as shadow baking.
That is because SwitchLight struggles to remove hard shadows as we demonstrate in Figure~\ref{Fig:exp:cmp_testset} and Figure~\ref{Fig:exp:cmp_switchlight}.
To address this, the concurrent work, WildCap, proposes first manually locating these artifacts and then adopting a specialized optimization-based method to explain the baking artifacts as local dark lighting effects.
With \emph{manual intervention}, WildCap demonstrates comparable delighting results to our \emph{fully automatic} method.
However, their region-based optimization leads to discontinuous artifacts as illustrated in Figure~\ref{Fig:exp:cmp_wildcap}. 
We emphasize that with a powerful delighting prior, we can fundamentally simplify the facial appearance capture problem to a large extent.

\begin{figure}[t]
    \centering
    \includegraphics[width=0.475\textwidth]{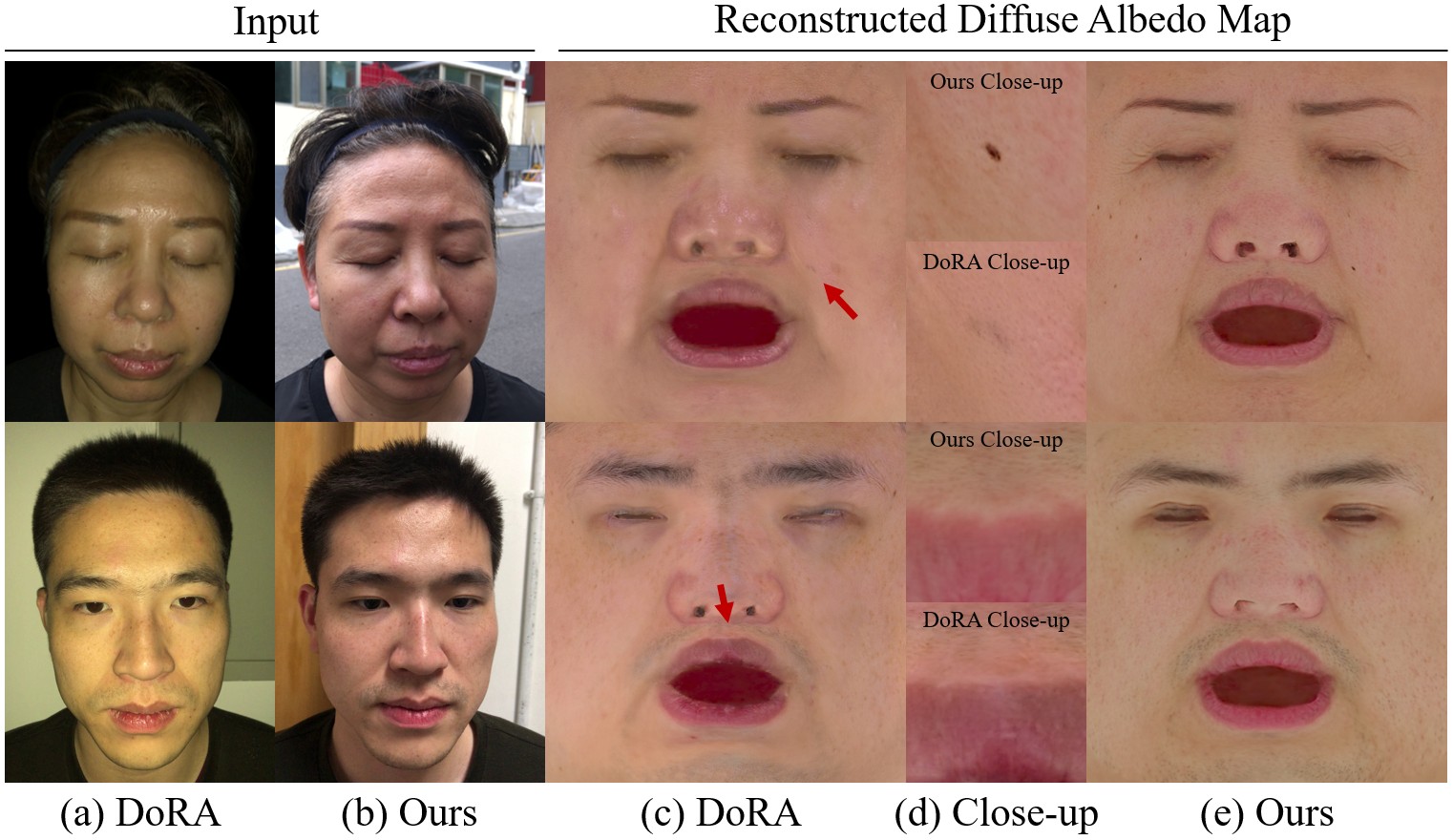}
    \caption{
    Comparison to DoRA on diffuse albedo reconstruction.
    We show an example captured image for both methods in (a) and (b), and results in (c), (d), and (e).
    }
    \label{Fig:exp:cmp_dora}
\end{figure}

\paragraph{Comparisons with DoRA}
DoRA~\cite{han2025dora} is a state-of-the-art method for facial appearance capture in controllable environments.
To compare with it, we capture an extra co-located smartphone and flashlight video for the same subject.
We feed the in-the-wild video to our method and the co-located video to DoRA.
As shown in Figure~\ref{Fig:exp:cmp_dora}, our method demonstrates comparable or better quality to DoRA while significantly reducing capture cost.
In addition, our method can better preserve person-specific facial traits such as nevus.
That is because our pipeline is built on top of WildCap~\cite{han2025wildcap}, which introduces an LPIPS loss term into the appearance capture process, while DoRA only adopts an L2 loss.
As geometry reconstruction and camera calibration are not perfectly accurate especially in our low-cost setup, using an L2 loss tends to average out facial details.

\begin{figure}[t]
    \centering
    \includegraphics[width=0.475\textwidth]{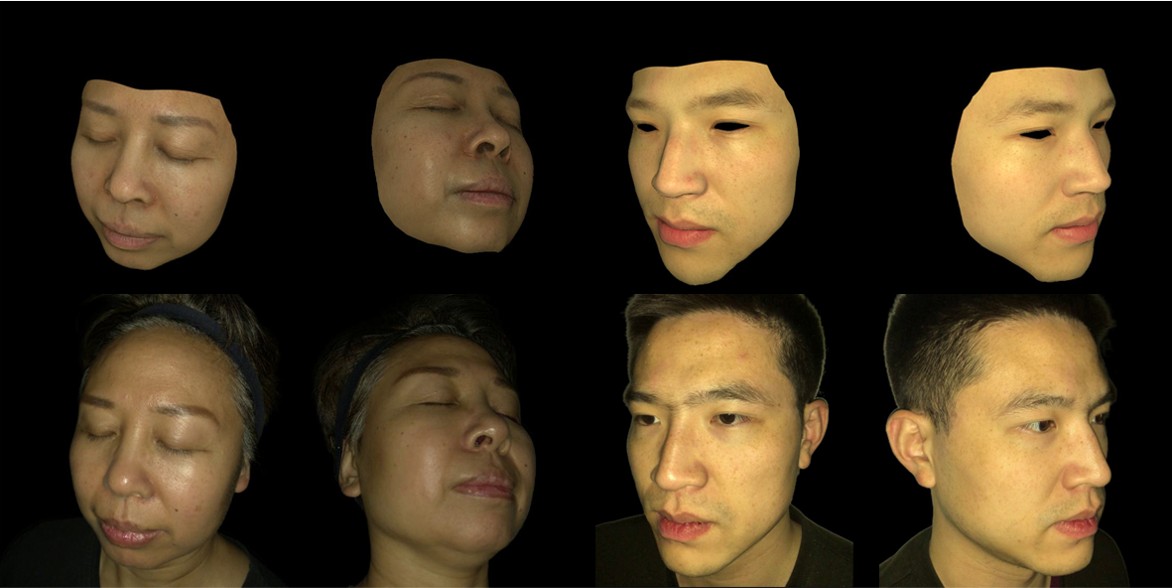}
    \caption{
    Evaluation on reconstruction fidelity of our facial appearance capture system. We show the re-rendered image and the real image in the first and second row, respectively. 
    }
    \label{Fig:exp:eval_fidelity}
\end{figure}

\begin{figure*}[p!]
    \centering
    \includegraphics[width=\textwidth]{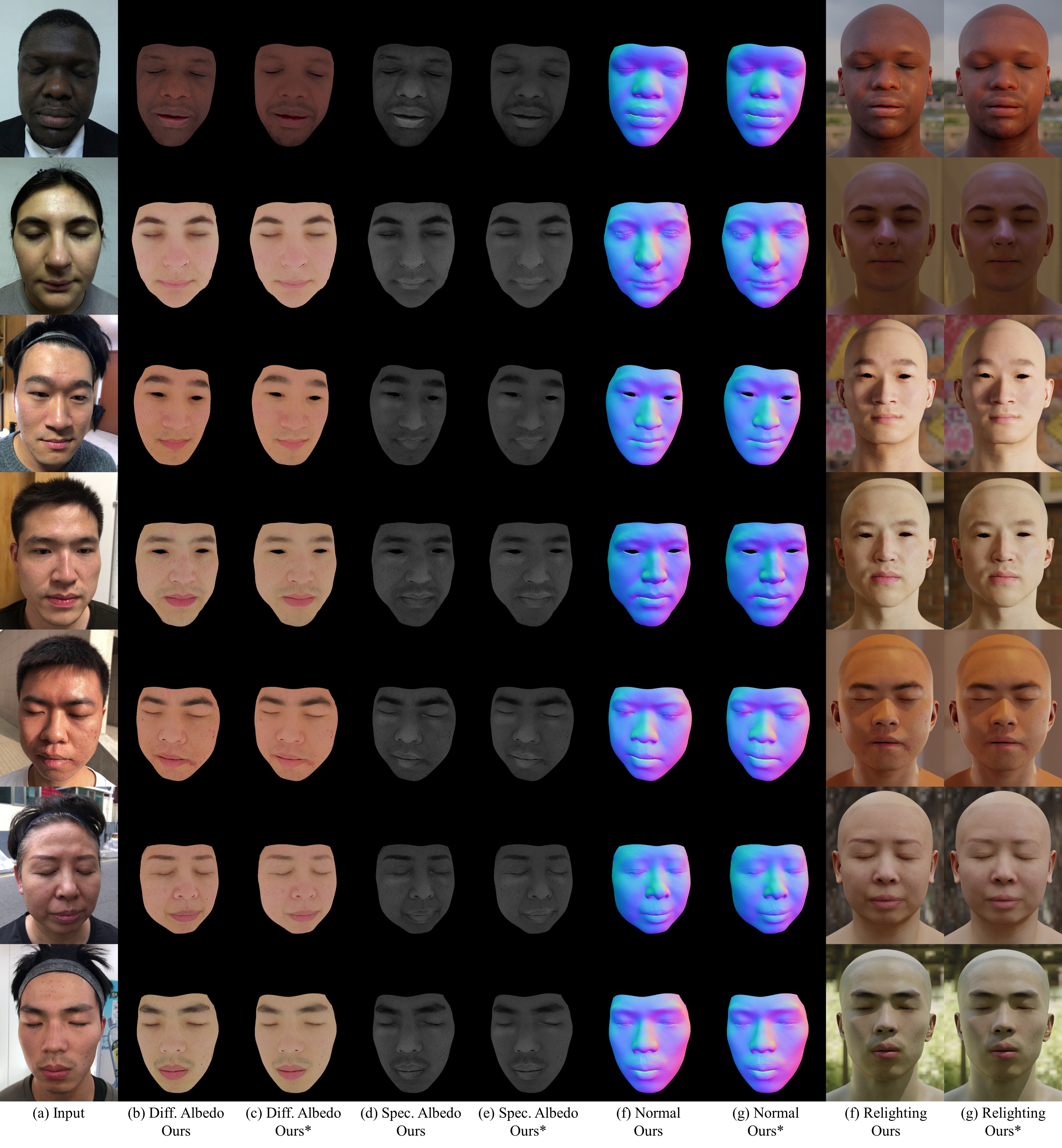}
    \caption{
    Facial appearance capture results of our method on subjects from diverse ethnic groups captured under complex in-the-wild environments.
    }
    \label{Fig:exp:more_results}
\end{figure*}

\subsubsection{Evaluation on Reconstruction Fidelity}\label{sec:exp:capture:recon_fidelity}
We evaluate the reconstruction fidelity of our method by comparing the re-rendered images with real-captured ones. 
To this end, we capture two sequences of the same subject: one in the wild and one in a controlled setup similar to DoRA~\cite{han2025dora}. 
We apply our method to the in-the-wild sequence and align the resulting geometry to the controlled setup for re-rendering under co-located lighting.
As shown in Figure~\ref{Fig:exp:eval_fidelity}, the re-rendered image looks close to the real image, demonstrating the high reconstruction fidelity of our method.
Regarding the quantitative metric, our method achieves an average LPIPS of 0.0576 compared to the real image. 
For context, the state-of-the-art method, DoRA~\cite{han2025dora}, achieves an LPIPS of 0.0499 using geometry reconstructed directly from the controlled sequence. 
Despite the increased difficulty of evaluating in-the-wild geometry against a separate controlled reference (whereas DoRA uses setup-specific reconstruction), our results remain highly competitive with DoRA.
In addition, this experiment also demonstrates the correctness of the reconstructed albedo of our method. 
Since the re-rendered results under a novel lighting setup closely match the ground-truth photograph, it provides empirical evidence that our estimated albedo does not bake in environment-specific lighting.

\subsubsection{More Results}\label{sec:exp:capture:more_results}

We test our method on subjects from diverse ethnic groups in Figure~\ref{Fig:exp:more_results}, where we show an example input image, the reconstructed reflectance maps rendered in screen space, and a relighting result.
As illustrated in Figure~\ref{Fig:exp:more_results}, both our method and the open-sourced version demonstrate plausible reflectance estimation results that largely close the quality gap between high-end setups.
In addition, due to the difference in training data, the skin tone in the raw predictions of OpenDelight and OpenDelight* are also different, as shown in Figure~\ref{Fig:exp:cmp_switchlight}.
Thanks to our prior-grounded optimization, the skin tone of \emph{Ours} and \emph{Ours*} are calibrated and become consistent. 
Please see our supplementary video for a better illustration.

\begin{figure}[t]
    \centering
    \includegraphics[width=0.475\textwidth]{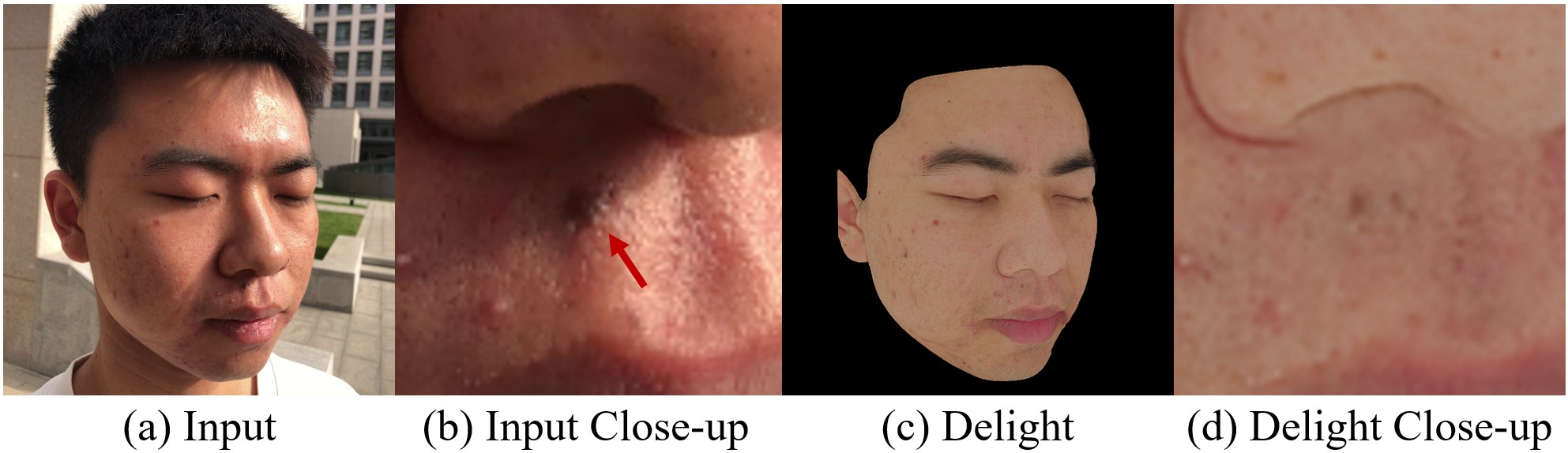}
    \caption{
    Limitations of our delighting network on processing colored nevus.
    We show the input image and its close-up in (a) and (b), and the predicted diffuse albedo and its close-up in (c) and (d).
    }
    \label{Fig:exp:limit_pore}
\end{figure}

\subsection{Discussions}\label{sec:exp:discuss}
\paragraph{Potential Identity Loss}
While our method achieves high-quality results, we observe potential identity loss arising from both the delighting and appearance capture processes.
Regarding the delighting process, since the training set of our delighting network has limited identities, we observe that our method sometimes tends to explain colored nevus as lighting effects and remove them from the diffuse albedo map, as shown in Figure~\ref{Fig:exp:limit_pore}. 
To address this, a possible solution is to train a relighting network together with the delighting network; this way, we can train on a large-scale single-view face dataset such as FFHQ~\cite{karras2019style} and enforce a self-supervised reconstruction loss similar to \citet{yeh2022learning}.
Regarding the appearance capture process, we also observe that the re-rendered 3D facial assets in Figure~\ref{Fig:exp:more_results} do not completely match the photograph. 
This discrepancy mainly arises from geometric inaccuracies in the WildCap system (Section~\ref{sec:inv_render:data}), particularly in template registration and full-head completion.

\paragraph{Skin Tone Bias}
Due to the unbalanced nature in the training dataset for both the delighting network and the diffusion prior (Section~\ref{sec:delight_prior:data} and ~\ref{sec:nsb}), we observe our method is more biased towards lighter skin tones; see the results on the African American subject in Figure~\ref{Fig:exp:more_results}.
We believe that collecting a more balanced dataset across diverse ethnic groups will be a valuable practical contribution.

\paragraph{Efficiency}
Our current method takes about 30 minutes from the smartphone video to the relightable 3D assets.
Following recent trends~\cite{Ming2025VGGTFaceTC,Wang2025VGGTVG}, learning a feed-forward network to directly map sampled frames to the relightable 3D facial assets is an interesting future work.



\section{Conclusion}
We propose a fully automatic system for capturing facial appearance from smartphone video recorded in the wild.
Considering that directly applying model-based inverse rendering is ill-posed and fragile to optimize, our key insight is to shift the focus of appearance capture to learning a powerful delighting prior.
Specifically, on the data front, we leverage two complementary datasets, an OLAT dataset and a rendered Light Stage scan dataset. 
To enable effective and efficient mix training, we propose a Dataset Latent Modulation technique where we distill the essential delighting mechanisms into the core network while using source-aware tokens to absorb other dataset-specific bias.
With simple network architecture and loss functions, we demonstrate superior delighting results to a leading proprietary method, SwitchLight.
We further demonstrate that our powerful delighting prior can significantly simplify the appearance capture process. 
Our automatic method achieves significantly better results than previous works, while comparable to a concurrent work that requires manual intervention.
Lastly, we leverage our method to transform the multi-view NeRSemble dataset into a large-scale collection of 4K-resolution relightable scans, which will be released with our code to foster future research on appearance capture and portrait delighting.

\begin{acks}
This work was supported by the NSFC (No.62561160115). This work was also supported by THUIBCS, Tsinghua University, and BLBCI, Beijing Municipal Education Commission. Feng Xu is the corresponding author.
\end{acks}

\bibliographystyle{ACM-Reference-Format}
\bibliography{sample-bibliography}

\end{document}